\documentclass{article}

\usepackage{microtype}
\usepackage{graphicx}
\usepackage{subcaption}
\usepackage{booktabs} % for professional tables
\usepackage{multirow}
\usepackage{pifont}
\usepackage[table,xcdraw]{xcolor}
\usepackage{hyperref}
\usepackage{enumitem}

\usepackage[preprint]{icml2026}

\usepackage{amsmath}
\usepackage{amssymb}
\usepackage{mathtools}
\usepackage{amsthm}
\usepackage{makecell}

\usepackage[capitalize,noabbrev]{cleveref}

\theoremstyle{plain}

\theoremstyle{definition}

\theoremstyle{remark}

\usepackage[textsize=tiny]{todonotes}

\icmltitlerunning{Discrete Prototypical Memories for Federated Time Series Foundation Models}

\begin{document}

\twocolumn[
  % \icmltitle{Revisiting Federated Foundation Models for Time Series: Towards a Unified Discrete Representation Space}
    \icmltitle{Discrete Prototypical Memories for Federated Time Series Foundation Models}



  \icmlsetsymbol{equal}{*}

  \begin{icmlauthorlist}
    \icmlauthor{Liwei Deng}{1,2}
    \icmlauthor{Qingxiang Liu}{2,equal}
    \icmlauthor{Xinhe Niu}{3}
    \icmlauthor{Shengchao Chen}{1}
    \icmlauthor{Sheng Sun}{4}
    \icmlauthor{Yuankai Wu}{5}
    \icmlauthor{Guodong Long}{1}
    \icmlauthor{Yuxuan Liang}{2}
  \end{icmlauthorlist}

  \icmlaffiliation{1}{Australian Artificial Intelligence Institute, University of Technology Sydney}
  \icmlaffiliation{2}{The Hong Kong University of Science and Technology (Guangzhou)}
  \icmlaffiliation{3}{Shenzhen University}
   \icmlaffiliation{4}{Institute of Computing Technology, Chinese Academy of Sciences, Beijing, China}
 \icmlaffiliation{5}{Sichuan University}
  \icmlcorrespondingauthor{Qingxiang Liu}{qingxiangliu737@gmail.com}
  \icmlcorrespondingauthor{Yuxuan Liang}{yuxliang@outlook.com}

  \icmlkeywords{Machine Learning, ICML}

  \vskip 0.3in
]



\printAffiliationsAndNotice{\icmlEqualContribution}

\begin{abstract}
Leveraging Large Language Models (LLMs) as federated learning (FL)–based time series foundation models offers a promising way to transfer the generalization capabilities of LLMs to time series data while preserving access to private data. However, the semantic misalignment between time-series data and the text-centric latent space of existing LLMs often leads to degraded performance. Meanwhile, the parameter-sharing mechanism in existing FL methods model heterogeneous cross-domain time-series data into unified continuous latent space, which contradicts to the fact that time-series semantics frequently manifest as discrete and recurring regimes.
To address these limitations, we propose \textsc{FeDPM}, a federated framework for time-series foundation models based on discrete prototypical memories. 
Specifically, we learn local prototypical memory priors for intra-domain time series data.
We then align cross-domain memories to promise the unified discrete latent space and introduce domain-specific memory update mechanism to balance shared and personalized prototypical knowledge.
Extensive experiments demonstrate the efficiency and effectiveness of \textsc{FeDPM}.
The code is publicly available at \url{https://anonymous.4open.science/r/FedUnit-64D1}.

% that \textsc{FeDPM} consistently achieves state-of-the-art performance, while reducing communication costs by up to 97.03\% and trainable parameters by over 20.37\%. 

\end{abstract}

\section{Introduction}
\begin{figure}[t]
    \centering
    \includegraphics[width=1\linewidth]{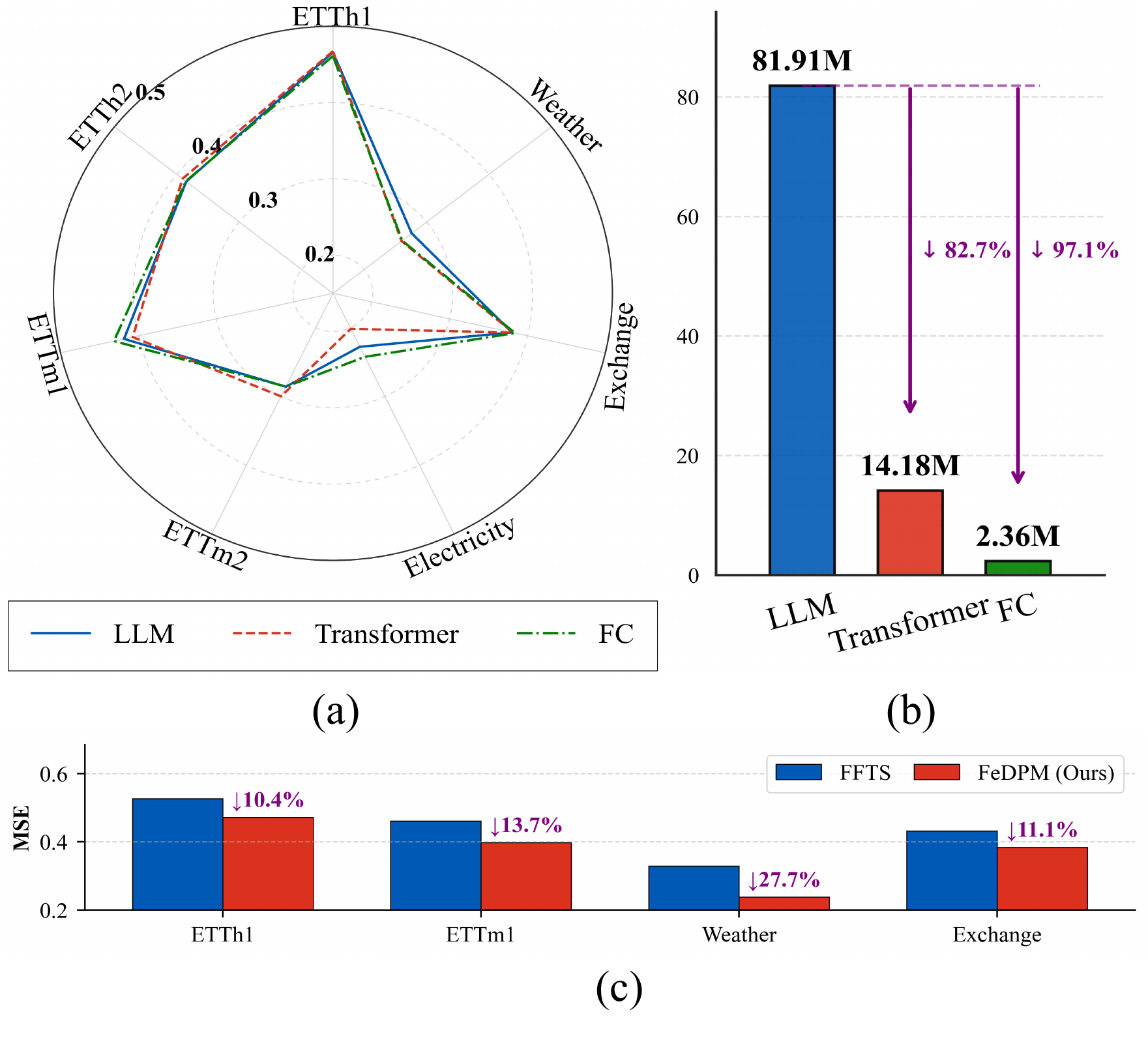}
    \caption{Ablation study of Time-FFM by replacing the frozen LLM backbone with trainable Transformer layers or FC layers on (a) forecasting MSE and (b) number of parameters. (Detailed settings and results in Appendix \ref{Ablation_Time-FFM}.) (c) Performance comparison between our proposed \textsc{FeDPM} and FFTS.}
    \label{fig:motivation}
\end{figure}

Time series forecasting plays a crucial role in a variety of real-world applications, such as energy consumption prediction \cite{time-vlm,D2Vformer}, weather forecasting \cite{airformer,STD2Vformer}, and disease transmission modeling \cite{liu2024moirai,time_series_survey}. Inspired by the remarkable success of Foundation Models (FMs) in natural language processing \cite{gpt3,deepseek-r1} and computer vision \cite{vit,kimi-vl}, there has been a surge of interest in developing general-purpose FMs for time series analysis \cite{time-llm,unitime,foundation_time_series_survey}. 
% However, existing FMs for time series are predominantly trained in a centralized manner, which requires aggregating data from multiple sources into a single repository. 
With the rapid scaling of FMs, the effectiveness of model performance increasingly follows established scaling laws \cite{scale_laws_LLM,scale_laws,scale_laws2}, which require ever-growing amounts of training data. However, most publicly available time series datasets are limited in scale and diversity, and are gradually being exhausted as model capacity continues to grow. This limitation motivates the exploitation of abundant private data distributed across different data owners. 

However, directly centralizing such data raises serious privacy concerns and may violate data protection regulations, such as the General Data Protection Regulation (GDPR)~\cite{gdpr} and California Consumer Privacy Act (CCPA)~\cite{CCPA}.
Federated Learning (FL) provides a promising paradigm for training FMs using the private data by merely exchaning intermediate model parameters. The recent studies have explored FL-based time series modeling by aligning temporal signals with the textual embedding space of pre-trained Large Language Models (LLMs)~\cite{time-ffm, FedTime, chen2023federated, chen2024personalized}. 
We conduct an ablation study on state-of-the-art Time-FFM to investigate \textit{whether pretrained LLMs can actually generalize to time series data in FL setting} (see Figure \ref{fig:motivation} (a) and (b)).PM
A key observation is that lightweight models achieve lower MSE in \textbf{71.43\%} of evaluation settings with only \textbf{10.1\%} parameters on average, which suggests a fundamental semantic misalignment between time series data and the text-centric latent space of existing LLMs. 
% Specifically, LLMs impose a rigid and pre-defined representation structure optimized for discrete tokens, which is ill-suited for modeling continuous numerical temporal patterns.

These findings motivate the need to construct representations that are native to time series dynamics. 
Most existing FL methods \cite{FeDaL,FFTS} rely on parameter-sharing mechanisms to transfer knowledge across domains by projecting heterogeneous time series into a unified continuous latent space. This design implicitly assuming that heterogeneous temporal patterns can be embedded into a unified continuous latent space without semantic distortion (see the prediction performance of FFTS in Figure~\ref{fig:motivation} (c)).
However, time series semantics often manifest as \textbf{discrete and recurring regimes}, such as the phase transitions in traffic flow (e.g., free-flow $\to$ synchronized $\to$ congested states), whose abrupt switches and non-smooth dynamics violate the smoothness assumption of continuous representations, potentially causing semantic entanglement and negative transfer in federated settings.

% an assumption that is particularly vulnerable to severe cross-domain data heterogeneity, including temporal resolution bias and physical constraint bias \cite{}. Moreover, these approaches predominantly model temporal signals in continuous latent spaces, even though time series semantics often emerge as discrete and recurring regimes (e.g., traffic patterns, workload phases, or physiological conditions).
% As a result, the misalignment between domain-specific biases and continuous representation spaces frequently leads to feature space collapse under non-IID data distributions   \cite{FL_Non_IID, FL_Non_IID2, FL_Non_IID_survey}.

To address these challenges, we propose \textsc{FeDPM}, a \textbf{Fe}derated framework for time series foundation model via \textbf{D}iscrete \textbf{P}rototypical \textbf{M}emories. 
% \textsc{FeDPM} constructs a foundation representation space for cross-domain time series data through a set of learnable prototypes that serve as universal latent memories. 
Specifically, each client \footnote{In this paper, we use ``client'' and ``domain'' interchangeably, as each client corresponds to a time series domain.} learns local prototypical memory priors that distill domain-specific temporal knowledge. Rather than exchanging full model parameters, clients and the server communicate only these prototypical memories. 
On the server side, we introduce the cross-domain memory update mechanism, which incorporates cross-domain memory alignment to promise the unified discrete latent space for cross-domain time series data and domain-specific memory update to balance the shared and personalized prototypical knowledge.
% that adaptively aggregates prototypical memories based on their semantic similarity, thereby forming a unified time-series representation space.
% Moreover, to balance shared knowledge and domain-specific personalization, we design a dual-component prototype mechanism:
% (1) Shared Prototypes act as universal semantic anchors that force heterogeneous domains to map similar temporal patterns into a unified coordinate system, enabling implicit feature alignment;
% (2) Personalized Prototypes serve as domain-specific complements, allowing each client to preserve unique local dynamics that would otherwise be lost in a purely shared representation space.
Our contributions are summarized as follows:
\begin{itemize}[leftmargin=*]
    \item \textbf{Conceptual}: We identify representation mismatch as a fundamental bottleneck for time series FMs under FL, highlighting the necessity of domain-native and unified discrete representations.
    \item \textbf{Methodological}: We propose \textsc{FeDPM}, a federated framework that introduces learnable discrete prototypical memories to balance shared and personalized knowledge, enabling effective semantic aggregation across heterogeneous domains without sharing raw data.
    \item \textbf{Practical}: We conduct extensive experiments on seven real-world benchmarks, where \textsc{FeDPM} consistently achieves state-of-the-art performance while reducing communication overhead by over 97.03\% and trainable parameters by over 20.37\% compared to existing FL baselines.
\end{itemize}

\section{Related Work}
\begin{figure*}[th]
    \centering
    \includegraphics[width=0.99\linewidth]{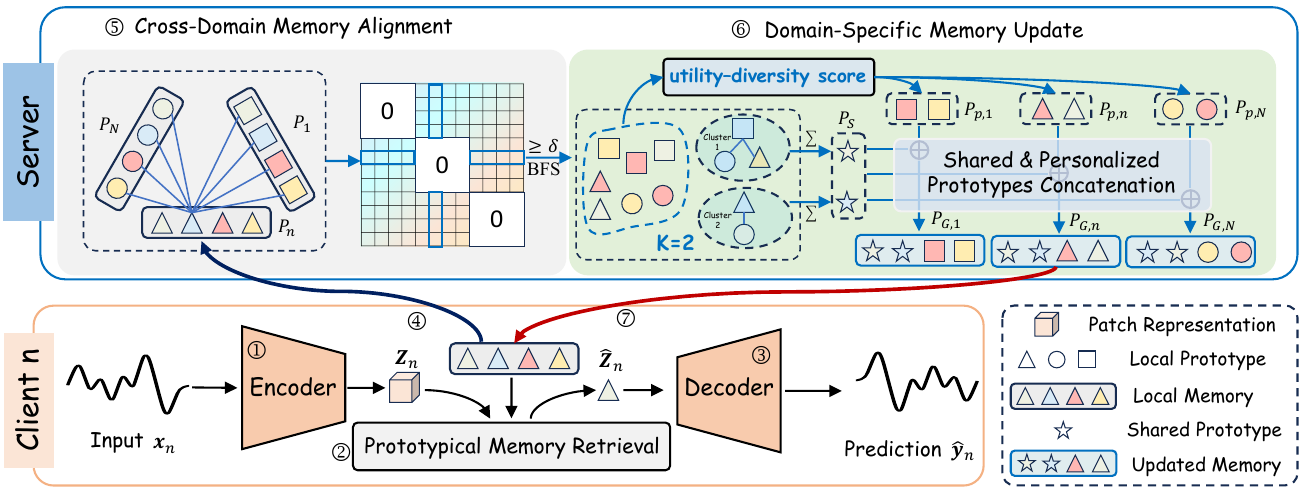}
    \caption{The overall architecture of \textsc{FeDPM}.}
    \label{Overview}
\end{figure*}

\paragraph{Foundation Models for Time Series Forecasting.}
% Pre-trained models have evolved into large FMs capable of handling diverse data modalities and tasks, exhibiting strong few-shot and zero-shot generalization abilities \cite{time-llm}. Inspired by this success, recent studies have begun to explore foundation models for time series forecasting (TSFMs). However, compared with vision and language domains, TSFMs remain relatively underexplored due to severe data heterogeneity and the lack of large-scale, high-quality time series corpora \cite{FeDaL}.

Existing efforts on foundation models for time series forecasting (TSFMs) can be broadly divided into two paradigms. One line of work adapts pretrained LLMs to time series forecasting by either fine-tuning a small subset of parameters \cite{GPT4TS,llm4ts} or reformulating time series into prompts or token sequences \cite{time-llm,unitime,tempo}. By treating time series as a modality-compatible input, these methods aim to exploit the general reasoning capabilities of LLMs, but their effectiveness heavily relies on the choice of backbone models and the quality of cross-modal alignment.
Another line of research focuses on training TSFMs from scratch using large-scale time series data \cite{forecastpfn,woo2024unified,TimeGPT-1,Moment,Timer}. Although these models demonstrate promising cross-domain generalization, they typically require substantial computational resources and centralized access to large-scale datasets, which limits their applicability in privacy-sensitive and distributed settings. Moreover, time series data are inherently heterogeneous across domains, sensors, and environments, and such heterogeneity further complicates model training and degrades forecasting accuracy in practice \cite{FFTS,federated_NonIID}.

\paragraph{Federated Learning in Time Series Forecasting.}
% FL provides a rich set of techniques to explicitly address such heterogeneity while preserving data privacy \cite{m2024personalized,jamali2022federated,long2023multi,Fedproto,federated_NonIID2}. This has motivated growing interest in integrating FL with time series forecasting, aiming to extend TSFMs from centralized settings to privacy-preserving and decentralized scenarios.

Existing studies on TSFMs under the FL paradigm largely follow the two modeling philosophies discussed above.
On the one hand, several works adapt pretrained  LLMs to federated time series forecasting by fine-tuning lightweight parameter subsets \cite{chen2024personalized} or constructing multimodal prompts to encode time series information \cite{time-ffm}. While these approaches reduce local training costs and leverage pretrained knowledge, they rely on the assumption that LLM backbones can faithfully capture time series dynamics. However, our empirical analysis (Figure~\ref{fig:motivation}), together with recent findings in \cite{tan2024language}, suggests that this assumption does not hold for current LLMs, especially under heterogeneous federated settings.
On the other hand, alternative approaches directly train TSFMs from scratch in a federated manner \cite{FFTS}. Although this line of work avoids dependence on LLM backbones, it typically requires frequent transmission of large model parameters, leading to substantial communication overhead. Moreover, parameter-based aggregation offers limited interpretability, making it difficult to understand how domain-specific temporal knowledge is transferred and integrated.
Taken together, these limitations underscore the need for communication-efficient and knowledge transfer mechanisms that are specifically designed for federated time series forecasting.

\section{Methodology}

% \subsection{Problem Definition}
Given $N$ domains, let $\mathcal{D}_n = \{(\pmb{X}_n, \pmb{Y}_n)\}$ denote the local dataset of domain $n$, 
% where $D_n = |\mathcal{D}_n|$ is the number of training samples. 
In the context of time series forecasting, we denote $\pmb{X}_n \in \mathbb{R}^{L_n\times c_n}$ as the input of the personalized prediction model $f_n(\cdot)$, where $L_n$ represents the domain-variant lookback window and $c_n$ represents the number of dimensions (channels). The ground truths can be denoted as  $\pmb{Y}_n \in \mathbb{R}^{F_n\times c_n}$, where $F_n$ represents the future prediction window. 
% Specifically, we denote the observation of domain $n$ at the time step $t$ as $\pmb{x}_{n,t}=\{x^1_{n,t},...,x^{c_n}_{n,t} \}\in \mathbb{R}^{c_n}$, where $c_n$ represents the number of dimensions (channels). 
For ease of reference, we summarize the commonly used notations in Table \ref{tab:notations} in the Appendix.

% \subsection{Model Structure}
Figure \ref{Overview} illustrates an overview of the proposed federated time series forecasting framework, termed \textsc{FeDPM}. Each client locally processes its private time series data using an \ding{172} encoder--\ding{174} decoder architecture, augmented with \ding{173} a \emph{Prototypical Memory Retrieval} module to access domain-specific prototypical memories.
To facilitate cross-domain knowledge sharing without exchanging raw data, each domain periodically \ding{175} uploads its locally learned memory $\pmb{P}_n$ to the server. The server then performs \ding{176} \emph{Cross-Domain Memory Alignment} to unify the discrete latent space and further performs \ding{177} \emph{Domain-Specific Memory Update}, deriving a set of shared prototypes $\pmb{P}_S$ that capture common temporal patterns, along with a set of personalized prototypes $\pmb{P}_{p,n}$ that preserve domain-specific information.
These two components are concatenated to form the global memory for domain $n$, denoted as $\pmb{P}_{G,n} = [\pmb{P}_S; \pmb{P}_{p,n}]$. The aggregated memory $\pmb{P}_{G,n}$ is subsequently \ding{178} transmitted back to the corresponding client and used to initialize the memory for the next round of local training.

\subsection{Local Prototypical Memory Priors}
% We describe the learning procedure of the local prototypical memory priors, which mainly consists of an Encoder–Decoder architecture augmented with a \textit{Prototypical Memory Retrieval} module.

\paragraph{Encoder Module.}
To accommodate domain-variant channels $c_n$, we adopt a channel-independent strategy \cite{patchtst} that processes each univariate time series, which is denoted as $\pmb{x}_n \in \mathbb{R}^{L_n}$ for simplicity.
% $\pmb{X}^j_n \in \mathbb{R}^{1\times L_n}$ separately. For simplicity, we omit the superscript $j$ and denote the series as $\pmb{x}_n \in \mathbb{R}^{L_n}$. 
Each series is first normalized by its instance-wise mean and standard deviation \cite{revin,san}, 
and then partitioned into non-overlapping patches of length $S_n$ with stride $S_n$, producing $B_n=\left\lceil\frac{L_n-S_n}{S_n}\right\rceil+1$ patches.
These patches, denoted as $\pmb{X}_{n,S}\in\mathbb{R}^{B_n\times S_n}$, are linearly projected into $D$-dimensional token embeddings $\hat{\pmb{X}}_{n,S}\in\mathbb{R}^{B_n\times D}$.
To model temporal dependencies in the patched sequence, we feed the token embeddings into a domain-specific encoder $\mathcal{M}_{n,\mathcal{E}}$. Our framework is agnostic to the architectural choice of $\mathcal{M}_{n,\mathcal{E}}$, and supports various instantiations (see Section \ref{sec:model_analysis}). The encoder outputs latent representations $\pmb{Z}_n \in \mathbb{R}^{B_n \times D}$.

\paragraph{Prototypical Memory Retrieval.}
To distill domain-specific knowledge from each domain while simultaneously incorporating information from other domains, we employ a Prototypical Memory Retrieval (PMR) mechanism as an effective medium for bridging local and global knowledge \cite{totem}.
% Specifically, a global memory $\pmb{P}_{G,n} \in \mathbb{R}^{M \times D}$ for domain $n$, which encodes shared semantic knowledge across all domains and personalized knowledge, is used to initialize the local memory of domain $n$, denoted as $\pmb{P}_n = \{\pmb{e}_{n,1}, \ldots, \pmb{e}_{n,M}\} \in \mathbb{R}^{M \times D}$.
% At the beginning of training, the global memories are randomly initialized and progressively refined through federated aggregation.
Specifically, given the encoder output $\pmb{Z}_n=\{\pmb{z}_{n,1},\ldots,\pmb{z}_{n,B_n}\}\in \mathbb{R}^{B_n \times D}$, we retrieve the most similar prototype for each patch-level latent representation $\pmb{z}_{n,b}\in \mathbb{R}^{D}$ by minimizing the Euclidean distance from  local memory of domain $n$, denoted as $\pmb{P}_n = \{\pmb{e}_{n,1}, \ldots, \pmb{e}_{n,M}\} \in \mathbb{R}^{M \times D}$:
\begin{align}
\hat{\pmb{z}}_{n,b} =\min_{1\le i\le M}||\pmb{z}_{n,b}-\pmb{e}_{n,i}||_2, \label{eq:vq_strat}
% sg(x) &= 
% \begin{cases} 
% x & (in \ forward \ propagation) \\
% 0 & (in \ backward \ propagation)
% \end{cases},
\end{align}
% where $\ell_{bn}$ denotes the index of the prototype that is closest to the patch $b_n$ latent representation $\pmb{z}_{n,b_n}$.
% The corresponding prototype $\pmb{e}_{\ell_{bn}} \in \mathbb{R}^{D}$ is then retrieved from the Memory according to the selected index.
% Since the $\arg\min(\cdot)$ operation is non-differentiable, we follow VQ-VAE \cite{vqvae} and adopt the straight-through gradient estimator implemented via the stop-gradient operator $sg(\cdot)$ to enable end-to-end optimization.
where $\hat{\pmb{z}}_{n,b} \in \mathbb{R}^{D}$ denotes the retrieved prototype and is termed as the patch-level quantized representation.
After applying PMR to all patches, the quantized representations are concatenated to form
$\hat{\pmb{Z}}_{n}= \{\hat{\pmb{z}}_{n,1}, \ldots, \hat{\pmb{z}}_{n,B_n}\}\in \mathbb{R}^{B_n \times D}$.
% In summary, PMR can be viewed as an analogue of temporal tokenization, which discretizes continuous time series into a finite set of semantic representations.
% By mapping the original continuous latent space to a discrete memory space, PMR effectively reduces the hypothesis space of the model, thereby simplifying the learning process.

\paragraph{Decoder Module.}
The decoder module recovers continuous temporal representations from the retrieved discrete prototypes. 
Given the PMR-processed latent representation $\hat{\pmb{Z}}_{n}$, We apply a domain-specific decoder $\mathcal{M}_{n,\mathcal{D}}$ to produce decoded representations $\hat{\pmb{H}}_n \in \mathbb{R}^{B_n \times D}$. 
To generate predictions aligned with the target horizon, the decoder outputs $\hat{\pmb{H}}_n$ are flattened and linearly projected into the target space, followed by a de-normalization layer to yield the final prediction $\hat{\pmb{y}}_n \in \mathbb{R}^{F_n}$.

% Similar to the encoder, our framework is architecture-agnostic to the specific design of the decoder $\mathcal{M}_{n,\mathcal{D}}$ for domain $n$. 
% Formally, the decoding process is defined as
% \begin{align}
%     \hat{\pmb{H}}^j_n=\mathcal{M}_{n,\mathcal{D}}(\hat{\pmb{Z}}^{j}_{n}),\label{eq:Decoder}
% \end{align}
% where $\hat{\pmb{H}}^j_n \in \mathbb{R}^{B_n \times D}$ denotes the decoder output.

% Since the input time series is normalized at the beginning to mitigate non-stationary effects, it is necessary to reintroduce these instance-specific non-stationary factors into the prediction.
% Following \cite{revin}, we employ a multi-layer perceptron (MLP) to estimate the variations of the instance-specific mean and standard deviation between the input and output sequences.
% Formally, this process is defined as:
% \begin{align}
%     (\Delta\mu^j_i \, \Vert \, \Delta\sigma^j_i) = \mathrm{MLP}(\mu^j_i \, \Vert \, \sigma^j_i),
% \end{align}
% where $\Vert$ denotes the concatenation operator, and $\Delta\mu^j_i$ and $\Delta\sigma^j_i$ represent the predicted changes of the instance-specific mean and standard deviation, respectively. After estimating the variations of the instance-specific mean and standard deviation, the de-normalization step can be formulated as:
% \begin{align}
%     \hat{\pmb{Y}}^j_i
%     = \overline{\pmb{Y}}^j_i \odot (\Delta\sigma^j_i + \sigma^j_i)
%     + (\Delta\mu^j_i + \mu^j_i),
% \end{align}
% where $\odot$ represents element-wise multiplication, and $\hat{\pmb{Y}}^j_i \in \mathbb{R}^{1 \times F_i}$ is the final forecasting result.

\subsection{Cross-Domain Memory Update}
% We introduce the Cross-Domain Memory Update, which aggregates domain-specific knowledge across heterogeneous domains to construct balanced shared and personalized semantic representations.
% The module consists of two key operations: \emph{Cross-Domain Memory Alignment} and \emph{Domain-Specific Memory Update}.

\paragraph{Cross-Domain Memory Alignment.}
A fundamental challenge in aligning cross-domain memories is that prototypes are inherently \textit{permutation-invariant}\footnote{Reordering prototypes within a memory does not affect retrieval results, analogous to attention mechanisms \cite{attention-based_permutation-invariant,boue2025deep}}. Consequently, typical federated aggregation methods that rely on index-wise correspondence   \cite{fedavg,FedProx} cannot be directly applied to memory aggregation.

To address this issue, we introduce a cross-domain memory alignment mechanism that aligns prototypes across domains based on semantic similarity prior to aggregation. 
% Specifically, we compute cross-domain cosine similarities between prototypes. 
Given the local memories of domains $m$ and $n$, denoted as $\pmb{P}_m = \{\pmb{e}_{m,1}, \dots, \pmb{e}_{m,M}\}$ and $\pmb{P}_n = \{\pmb{e}_{n,1}, \dots, \pmb{e}_{n,M}\}$, the cosine similarity between the $i$-th prototype of domain $m$ and the $j$-th prototype of domain $n$ $(m \neq n)$ is defined as:
\begin{align}
    s^{m,n}_{i,j}
    =
    \frac{\pmb{e}_{m,i}^\top \pmb{e}_{n,j}}
    {\lVert \pmb{e}_{m,i} \rVert_2 \, \lVert \pmb{e}_{n,j} \rVert_2}. \label{eq:cos}
\end{align}
% where $\pmb{e}_{m,i}, \pmb{e}_{n,j} \in \mathbb{R}^{D}$ denote the corresponding prototypes from domains $m$ and $n$.
The resulting similarity matrix $\pmb{\mathcal{S}}^{m,n} = \{s^{m,n}_{i,j} \} \in \mathbb{R}^{M \times M}$ captures cross-domain prototype-wise semantic correlation. Prototype pairs with similarity scores exceeding a threshold $\delta$ are connected by undirected edges, forming a graph over  prototypes for different domains.
We identify semantic clusters by extracting the connected components of this graph using Breadth-First Search (BFS) \cite{BFS}. Each connected component corresponds to a cluster of semantically aligned prototypes across different domains. Let $\mathcal{K} = \{ \mathcal{I}_1, \dots, \mathcal{I}_{|\mathcal{K}|} \}$ denote the resulting set of clusters, where $\mathcal{I}_s$ contains the prototypes in the $s$-th cluster.

\paragraph{Domain-Specific Memory Update.}
Based on the semantic clustering results $\mathcal{K}$, we derive a shared representative prototype for each cluster by aggregating its constituent prototypes via mean pooling:
\begin{align} 
\pmb{e}_s = \frac{1}{|\mathcal{I}_s|} \sum_{\pmb{e}_i \in \mathcal{I}_s} \pmb{e}_i, \quad s = 1, \dots, |\mathcal{K}|, \label{eq:cluster}
\end{align}
where $\pmb{e}_i$ denotes the $i$-th prototype contributed by different domains, and $|\mathcal{I}_s|$ represents the cluster size. The resulting $\pmb{e}_s$ captures domain-shared semantic knowledge within the $s$-th cluster.

To balance globally shared knowledge with domain-specific nuances, we explicitly constrain the proportion of global prototypes in the memory.
Specifically, the number of shared prototypes is limited to at most a fraction $\gamma$ of the total memory size $M$, resulting in a maximum global capacity of $M_g=\lfloor \gamma M \rfloor$.
To prioritize global consensus while preserving personalization, we select the top-$K$ clusters with the largest cardinality, where $K=\min(|\mathcal{K}|, M_g)$. The centroids of these clusters are used to construct the shared prototypes $\pmb{P}_S \in \mathbb{R}^{K \times D}$, which captures semantic patterns consistently shared across domains.
The remaining memory size $M-K$, 
% accounting for at least $(1-\gamma)$ of the memory capacity, 
is reserved for domain-specific representations.

For each domain $n$, we construct personalized prototypes $\pmb{P}_{p,n}\in\mathbb{R}^{(M-K)\times D}$ by selecting prototypes from the unclustered set  $\mathcal{U}_n$.
This selection is guided by a \textit{utility–diversity score}, which favors informative yet non-redundant domain-specific patterns.
Given the $j$-th prototype of domain $n$, $\pmb{e}_{n,j}\in \mathcal{U}_n$, we obtain the score as:
\begin{equation}
\mathcal{V}(\pmb{e}_{n,j}) =
\frac{\mathrm{Freq}(\pmb{e}_{n,j})}{ \max_{\pmb{e} \in \mathcal{U}_n} \mathrm{Freq}(\pmb{e})}- \max_{\pmb{e} \in \mathcal{U}_{\text{other}}} \mathrm{Sim}(\pmb{e}_{n,j}, \pmb{e}),\label{eq:score}
\end{equation}
where $\mathrm{Freq}(\pmb{e}_{n,j})$ denotes the total number of patch-level representations assigned to prototype $\pmb{e}_{n,j}$ over one epoch of local training. This term favors reliable and informative prototypes, while down-weighting poorly trained or noisy ones. In addition, $\mathrm{Sim}(\cdot,\cdot)$ represents the cosine similarity defined in Eq. \eqref{eq:cos}, which explicitly penalizes high similarity between prototypes from different domains, thereby enhancing the preservation of domain-specific personalized knowledge.
Here, $\mathcal{U}_n$ denotes the unclustered prototypes of domain $n$, while $\mathcal{U}_{\text{other}}$ represents the union of unclustered prototypes from all other domains.
% This scoring function favors prototypes that are both frequently used within a domain and semantically distinct from other domains, thereby preserving personalized knowledge while suppressing noisy or redundant representations. 
Finally, we construct the domain-specific global memory by concatenating the shared prototypes $\pmb{P}_S$ and the personalized prototypes $\pmb{P}_{p,n}$, yielding $\pmb{P}_{G,n}=[\pmb{P}_S;\pmb{P}_{p,n}] \in \mathbb{R}^{M \times D}$ for domain $n$.

\begin{table*}[t]
\small
\centering
\tabcolsep=0.08cm
\renewcommand\arraystretch{1}
\caption{Comparison of \textsc{FeDPM} with representative Time-FFM and FFTS.}
\label{tab:method_comparison}
\begin{tabular*}{\textwidth}{c @{\extracolsep{\fill}}cccccc}
\toprule
\textbf{Method} & \textbf{Latent Space} & \textbf{Limitation} & \textbf{Comm. Object} & \textbf{Comm. Efficiency}  & \textbf{FM Construction} & \textbf{Params} \\
\midrule
\textbf{Time-FFM} & Text-centric & Semantic Misalignment & Prompts / Params & Low  & Stacking Params & High \\
\textbf{FFTS} & Continuous & Feature Collapse & Model Params & Low  & Stacking Params & High \\
\midrule
\textbf{\textsc{FeDPM}} & \textbf{Discrete Prototype} & \textbf{---} & \textbf{Memory Only} & \textbf{High} & \textbf{Unified Memory} & \textbf{Low} \\
\bottomrule
\end{tabular*}%
\end{table*}

\subsection{Training \& Inference}
\paragraph{Training.} To jointly optimize all trainable components of the proposed framework, we formulate a multi-term training objective.
Since the loss formulation is shared across all domains and channels, we focus on a channel of domain $n$ as a representative case. For notational consistency with the methodology, we directly adopt the previously defined variables, which simplifies the exposition without loss of generality. The overall objective is formulated as:
\begin{align}
\mathcal{L} &= \mathcal{L}_{\text{Pred}} + \beta \mathcal{L}_{\mathcal{M_E}} + \mathcal{L}_{\mathcal{M_C}}, \\
\mathcal{L}_{\text{Pred}} &= \text{Smooth}_{\text{L1}}(\hat{\pmb{y}}_n, \pmb{y}_n), \\
\mathcal{L}_{\mathcal{M_E}} &= || \pmb{Z}_n - sg(\hat{\pmb{Z}}_{n}) ||_2^2, \\
\mathcal{L}_{\mathcal{M_C}} &= || sg(\pmb{Z}_n) -\hat{\pmb{Z}}_{n} ||_2^2 ,
\end{align}
where $\pmb{y}_n \in \mathbb{R}^{F_n}$ denotes the ground-truth forecasting target, and $\text{Smooth}_{\text{L1}}(\cdot)$ is the Smooth L1 loss \cite{smoothl1,huber1992robust}, which improves robustness to outliers commonly observed in time series data \cite{totem}. Specifically, the decoder optimises only the first loss term, the encoder jointly optimises the first and second loss terms, while the prototypical memories are updated solely through the last loss term.
To enable effective learning of the discrete memory, we adopt the PMR objective from VQ-VAE \cite{vqvae}, where $sg(\cdot)$ denotes the stop-gradient operator. 
For completeness, the overall procedure is summarized in Algorithm \ref{alg:fl_vqvae} in Appendix \ref{appendix:Training_Process}.

\paragraph{Inference.} A domain-specific global memory is obtained for each domain and download to the corresponding client. During inference, inference data are processed locally by a domain-specific encoder--decoder architecture augmented with the PMR module to produce predictions.

\begin{table*}[th]
\centering
\setlength{\tabcolsep}{3pt}  % 减少列间距(默认6pt)
\renewcommand{\arraystretch}{1}  % 压缩行高
\caption{Full forecasting performance comparison results. \textbf{Bold} highlights the best performance across all methods, while \colorbox[HTML]{DDEBF7}{Blue} marks the best result among FL-FMs. ``Comm. Params.'' denotes the number of communicated parameters.
} 
\label{tab:Full-shot Result}
\resizebox{2\columnwidth}{!}{%
\begin{tabular}{cc|cc|cc|cc|cc|cc|cc|cc|cc|cc|cc|cc|cc|cc}
\toprule
\multicolumn{2}{c|}{Type}               & \multicolumn{10}{c|}{FL-FM}                                                                                                                                                                                                                                                                                                                                                  & \multicolumn{6}{c|}{Cen-FM}                                                                                                                                                                                         & \multicolumn{10}{c}{Expert}                                                                                                                                                                                                                                                                                        \\
\midrule
\multicolumn{2}{c|}{Method}             & \multicolumn{2}{c|}{\textsc{FeDPM}}                                                        & \multicolumn{2}{c|}{Time-FFM}                                           & \multicolumn{2}{c|}{FFTS}                                                        & \multicolumn{2}{c|}{FL-iTransformer}                                             & \multicolumn{2}{c|}{FL-PatchTST}              & \multicolumn{2}{c|}{TOTEM}                                            & \multicolumn{2}{c|}{UniTime}                                                   & \multicolumn{2}{c|}{Cen-PatchTST}                            & \multicolumn{2}{c|}{TimeNet}                                                   & \multicolumn{2}{c|}{Dlinear}                                                   & \multicolumn{2}{c|}{FEDformer}                                        & \multicolumn{2}{c|}{iTransformer}             & \multicolumn{2}{c}{PatchTST} \\
\midrule
\multicolumn{2}{c|}{Metric}             & MSE                                    & MAE                                    & MSE                           & MAE                                    & MSE                                    & MAE                                    & MSE                                    & MAE                                    & MSE                   & MAE                  & MSE                          & MAE                                   & MSE                                   & MAE                                   & MSE                          & MAE                          & MSE                                   & MAE                                   & MSE                                   & MAE                                   & MSE                                   & MAE                          & MSE                   & MAE                  & MSE      & MAE               \\
\midrule
                                & 96   & \cellcolor[HTML]{DDEBF7}0.391          & \cellcolor[HTML]{DDEBF7}0.407          & 0.406                         & 0.411                                  & 0.417                                  & 0.445                                  & 0.473                                  & 0.453                                  & 0.459                 & 0.457                & 0.402                        & 0.405                                 & {\color[HTML]{1F1F1F} 0.397}          & {\color[HTML]{1F1F1F} 0.418}          & {\color[HTML]{1F1F1F} 0.433} & {\color[HTML]{1F1F1F} 0.422} & {\color[HTML]{1F1F1F} 0.384}          & {\color[HTML]{1F1F1F} 0.402}          & {\color[HTML]{1F1F1F} 0.386}          & {\color[HTML]{1F1F1F} \textbf{0.400}} & {\color[HTML]{1F1F1F} \textbf{0.376}} & {\color[HTML]{1F1F1F} 0.419} & 0.387                 & 0.405                & 0.414    & 0.419             \\
                                & 192  & \cellcolor[HTML]{DDEBF7}0.441          & \cellcolor[HTML]{DDEBF7}0.434          & 0.460                         & 0.442                                  & 0.475                                  & 0.487                                  & 0.504                                  & 0.476                                  & 0.491                 & 0.474                & {\color[HTML]{1F1F1F} 0.457} & {\color[HTML]{1F1F1F} 0.436}          & {\color[HTML]{1F1F1F} 0.434}          & {\color[HTML]{1F1F1F} 0.439}          & {\color[HTML]{1F1F1F} 0.467} & {\color[HTML]{1F1F1F} 0.444} & {\color[HTML]{1F1F1F} 0.436}          & {\color[HTML]{1F1F1F} \textbf{0.429}} & {\color[HTML]{1F1F1F} 0.437}          & {\color[HTML]{1F1F1F} 0.432}          & {\color[HTML]{1F1F1F} \textbf{0.420}} & {\color[HTML]{1F1F1F} 0.448} & 0.441                 & 0.436                & 0.460    & 0.445             \\
                                & 336  & \cellcolor[HTML]{DDEBF7}0.486          & 0.463                                  & 0.504                         & \cellcolor[HTML]{DDEBF7}\textbf{0.453} & 0.531                                  & 0.521                                  & 0.535                                  & 0.494                                  & 0.549                 & 0.507                & 0.498                        & 0.461                                 & {\color[HTML]{1F1F1F} 0.470}          & {\color[HTML]{1F1F1F} 0.457}          & {\color[HTML]{1F1F1F} 0.509} & 0.472                        & {\color[HTML]{1F1F1F} 0.491}          & {\color[HTML]{1F1F1F} 0.469}          & {\color[HTML]{1F1F1F} 0.481}          & {\color[HTML]{1F1F1F} 0.459}          & {\color[HTML]{1F1F1F} \textbf{0.459}} & {\color[HTML]{1F1F1F} 0.465} & 0.491                 & 0.462                & 0.501    & 0.466             \\
\multirow{-4}{*}{\rotatebox{90}{ETTh1}}         & 720  & 0.572                                  & 0.508                                  & \cellcolor[HTML]{DDEBF7}0.495 & \cellcolor[HTML]{DDEBF7}\textbf{0.466} & 0.686                                  & 0.611                                  & 0.572                                  & 0.524                                  & 0.577                 & 0.526                & 0.539                        & 0.513                                 & {\color[HTML]{1F1F1F} 0.472}          & {\color[HTML]{1F1F1F} 0.477}          & {\color[HTML]{1F1F1F} 0.503} & 0.485                        & {\color[HTML]{1F1F1F} 0.521}          & {\color[HTML]{1F1F1F} 0.500}          & {\color[HTML]{1F1F1F} 0.519}          & {\color[HTML]{1F1F1F} 0.516}          & {\color[HTML]{1F1F1F} 0.506}          & {\color[HTML]{1F1F1F} 0.507} & 0.509                 & 0.494                & 0.496    & 0.481             \\
 \midrule
                                & 96   & 0.304                                  & \cellcolor[HTML]{DDEBF7}\textbf{0.343} & {\color[HTML]{1F1F1F} 0.305}  & {\color[HTML]{1F1F1F} 0.351}           & \cellcolor[HTML]{DDEBF7}\textbf{0.275} & 0.367                                  & 0.360                                  & 0.378                                  & 0.306                 & 0.353                & {\color[HTML]{1F1F1F} 0.299} & {\color[HTML]{1F1F1F} 0.343}          & {\color[HTML]{1F1F1F} 0.296}          & {\color[HTML]{1F1F1F} 0.345}          & {\color[HTML]{1F1F1F} 0.314} & {\color[HTML]{1F1F1F} 0.361} & {\color[HTML]{1F1F1F} 0.353}          & {\color[HTML]{1F1F1F} 0.374}          & {\color[HTML]{1F1F1F} 0.333}          & {\color[HTML]{1F1F1F} 0.387}          & {\color[HTML]{1F1F1F} 0.358}          & {\color[HTML]{1F1F1F} 0.397} & 0.301                 & 0.350                & 0.312    & 0.360             \\
                                & 192  & 0.377                                  & 0.392                                  & {\color[HTML]{1F1F1F} 0.380}  & {\color[HTML]{1F1F1F} 0.397}           & \cellcolor[HTML]{DDEBF7}\textbf{0.303} & \cellcolor[HTML]{DDEBF7}\textbf{0.385} & 0.447                                  & 0.434                                  & 0.392                 & 0.402                & {\color[HTML]{1F1F1F} 0.389} & {\color[HTML]{1F1F1F} 0.395}          & {\color[HTML]{1F1F1F} 0.374}          & {\color[HTML]{1F1F1F} 0.394}          & {\color[HTML]{1F1F1F} 0.407} & {\color[HTML]{1F1F1F} 0.411} & {\color[HTML]{1F1F1F} 0.402}          & {\color[HTML]{1F1F1F} 0.414}          & {\color[HTML]{1F1F1F} 0.477}          & {\color[HTML]{1F1F1F} 0.476}          & {\color[HTML]{1F1F1F} 0.429}          & {\color[HTML]{1F1F1F} 0.439} & 0.380                 & 0.400                & 0.388    & 0.405             \\
                                & 336  & 0.426                                  & 0.433                                  & {\color[HTML]{1F1F1F} 0.428}  & {\color[HTML]{1F1F1F} 0.436}           & \cellcolor[HTML]{DDEBF7}\textbf{0.328} & \cellcolor[HTML]{DDEBF7}\textbf{0.401} & 0.492                                  & 0.467                                  & 0.427                 & 0.435                & {\color[HTML]{1F1F1F} 0.448} & {\color[HTML]{1F1F1F} 0.436}          & {\color[HTML]{1F1F1F} 0.415}          & {\color[HTML]{1F1F1F} 0.427}          & {\color[HTML]{1F1F1F} 0.437} & {\color[HTML]{1F1F1F} 0.443} & {\color[HTML]{1F1F1F} 0.452}          & {\color[HTML]{1F1F1F} 0.452}          & {\color[HTML]{1F1F1F} 0.594}          & {\color[HTML]{1F1F1F} 0.541}          & {\color[HTML]{1F1F1F} 0.496}          & {\color[HTML]{1F1F1F} 0.487} & 0.428                 & 0.432                & 0.426    & 0.437             \\
\multirow{-4}{*}{\rotatebox{90}{ETTh2}}         & 720  & 0.555                                  & 0.530                                  & {\color[HTML]{1F1F1F} 0.427}  & {\color[HTML]{1F1F1F} 0.445}           & \cellcolor[HTML]{DDEBF7}\textbf{0.384} & \cellcolor[HTML]{DDEBF7}\textbf{0.434} & 0.539                                  & 0.500                                  & 0.448                 & 0.458                & {\color[HTML]{1F1F1F} 0.610} & {\color[HTML]{1F1F1F} 0.567}          & {\color[HTML]{1F1F1F} 0.425}          & {\color[HTML]{1F1F1F} 0.444}          & {\color[HTML]{1F1F1F} 0.434} & {\color[HTML]{1F1F1F} 0.448} & {\color[HTML]{1F1F1F} 0.462}          & {\color[HTML]{1F1F1F} 0.468}          & {\color[HTML]{1F1F1F} 0.831}          & {\color[HTML]{1F1F1F} 0.657}          & {\color[HTML]{1F1F1F} 0.463}          & {\color[HTML]{1F1F1F} 0.474} & 0.430                 & 0.447                & 0.433    & 0.453             \\
 \midrule
                                & 96   & \cellcolor[HTML]{DDEBF7}\textbf{0.324} & \cellcolor[HTML]{DDEBF7}\textbf{0.359} & 0.357                         & 0.373                                  & 0.380                                  & 0.405                                  & 0.379                                  & 0.389                                  & 0.647                 & 0.511                & {\color[HTML]{1F1F1F} 0.380} & {\color[HTML]{1F1F1F} 0.392}          & {\color[HTML]{1F1F1F} 0.339}          & {\color[HTML]{1F1F1F} 0.378}          & {\color[HTML]{1F1F1F} 0.927} & {\color[HTML]{1F1F1F} 0.604} & {\color[HTML]{1F1F1F} 0.338}          & {\color[HTML]{1F1F1F} 0.375}          & {\color[HTML]{1F1F1F} 0.345}          & {\color[HTML]{1F1F1F} 0.372}          & {\color[HTML]{1F1F1F} 0.379}          & {\color[HTML]{1F1F1F} 0.419} & 0.342                 & 0.377                & 0.344    & 0.373             \\
                                & 192  & \cellcolor[HTML]{DDEBF7}\textbf{0.382} & \cellcolor[HTML]{DDEBF7}0.392          & 0.399                         & 0.393                                  & 0.435                                  & 0.436                                  & 0.438                                  & 0.423                                  & 0.666                 & 0.516                & {\color[HTML]{1F1F1F} 0.406} & {\color[HTML]{1F1F1F} 0.403}          & {\color[HTML]{1F1F1F} 0.384}          & {\color[HTML]{1F1F1F} 0.403}          & {\color[HTML]{1F1F1F} 0.964} & {\color[HTML]{1F1F1F} 0.620} & {\color[HTML]{1F1F1F} 0.374}          & {\color[HTML]{1F1F1F} 0.387}          & {\color[HTML]{1F1F1F} 0.380}          & {\color[HTML]{1F1F1F} 0.389}          & {\color[HTML]{1F1F1F} 0.426}          & {\color[HTML]{1F1F1F} 0.441} & 0.383                 & 0.396                & 0.367    & \textbf{0.386}    \\
                                & 336  & \cellcolor[HTML]{DDEBF7}\textbf{0.409} & \cellcolor[HTML]{DDEBF7}\textbf{0.410} & 0.428                         & 0.417                                  & 0.485                                  & 0.470                                  & 0.504                                  & 0.460                                  & 0.685                 & 0.534                & {\color[HTML]{1F1F1F} 0.432} & {\color[HTML]{1F1F1F} 0.423}          & {\color[HTML]{1F1F1F} 0.412}          & {\color[HTML]{1F1F1F} 0.422}          & {\color[HTML]{1F1F1F} 1.041} & {\color[HTML]{1F1F1F} 0.656} & {\color[HTML]{1F1F1F} 0.410}          & {\color[HTML]{1F1F1F} 0.411}          & {\color[HTML]{1F1F1F} 0.413}          & {\color[HTML]{1F1F1F} 0.413}          & {\color[HTML]{1F1F1F} 0.445}          & {\color[HTML]{1F1F1F} 0.459} & 0.426                 & 0.420                & 0.399    & \textbf{0.410}    \\
\multirow{-4}{*}{\rotatebox{90}{ETTm1}}         & 720  & \cellcolor[HTML]{DDEBF7}0.475          & 0.461                                  & 0.490                         & \cellcolor[HTML]{DDEBF7}0.444          & 0.543                                  & 0.518                                  & 0.579                                  & 0.499                                  & 0.683                 & 0.557                & {\color[HTML]{1F1F1F} 0.497} & {\color[HTML]{1F1F1F} 0.471}          & {\color[HTML]{1F1F1F} 0.466}          & {\color[HTML]{1F1F1F} 0.451}          & {\color[HTML]{1F1F1F} 0.950} & {\color[HTML]{1F1F1F} 0.636} & {\color[HTML]{1F1F1F} \textbf{0.410}} & {\color[HTML]{1F1F1F} 0.450}          & {\color[HTML]{1F1F1F} 0.474}          & {\color[HTML]{1F1F1F} 0.453}          & {\color[HTML]{1F1F1F} 0.543}          & {\color[HTML]{1F1F1F} 0.490} & 0.491                 & 0.460                & 0.464    & \textbf{0.442}    \\
 \midrule
                                & 96   & \cellcolor[HTML]{DDEBF7}\textbf{0.178} & \cellcolor[HTML]{DDEBF7}\textbf{0.255} & {\color[HTML]{1F1F1F} 0.181}  & {\color[HTML]{1F1F1F} 0.267}           & 0.185                                  & 0.302                                  & 0.212                                  & 0.277                                  & 0.195                 & 0.282                & {\color[HTML]{1F1F1F} 0.197} & {\color[HTML]{1F1F1F} 0.274}          & {\color[HTML]{1F1F1F} 0.183}          & {\color[HTML]{1F1F1F} 0.266}          & {\color[HTML]{1F1F1F} 0.240} & {\color[HTML]{1F1F1F} 0.318} & {\color[HTML]{1F1F1F} 0.187}          & {\color[HTML]{1F1F1F} 0.267}          & {\color[HTML]{1F1F1F} 0.193}          & {\color[HTML]{1F1F1F} 0.292}          & {\color[HTML]{1F1F1F} 0.203}          & {\color[HTML]{1F1F1F} 0.287} & 0.186                 & 0.272                & 0.177    & 0.260             \\
                                & 192  & 0.253                                  & \cellcolor[HTML]{DDEBF7}0.307          & {\color[HTML]{1F1F1F} 0.247}  & {\color[HTML]{1F1F1F} 0.311}           & \cellcolor[HTML]{DDEBF7}\textbf{0.205} & 0.317                                  & 0.282                                  & 0.325                                  & 0.262                 & 0.318                & {\color[HTML]{1F1F1F} 0.258} & {\color[HTML]{1F1F1F} 0.315}          & {\color[HTML]{1F1F1F} 0.251}          & {\color[HTML]{1F1F1F} 0.310}          & {\color[HTML]{1F1F1F} 0.301} & {\color[HTML]{1F1F1F} 0.352} & {\color[HTML]{1F1F1F} 0.249}          & {\color[HTML]{1F1F1F} 0.309}          & {\color[HTML]{1F1F1F} 0.284}          & {\color[HTML]{1F1F1F} 0.362}          & {\color[HTML]{1F1F1F} 0.269}          & {\color[HTML]{1F1F1F} 0.328} & 0.254                 & 0.314                & 0.246    & \textbf{0.305}    \\
                                & 336  & 0.336                                  & \cellcolor[HTML]{DDEBF7}\textbf{0.289} & {\color[HTML]{1F1F1F} 0.309}  & {\color[HTML]{1F1F1F} 0.347}           & \cellcolor[HTML]{DDEBF7}\textbf{0.235} & 0.338                                  & 0.351                                  & 0.372                                  & 0.320                 & 0.353                & {\color[HTML]{1F1F1F} 0.330} & {\color[HTML]{1F1F1F} 0.363}          & {\color[HTML]{1F1F1F} 0.319}          & {\color[HTML]{1F1F1F} 0.351}          & {\color[HTML]{1F1F1F} 0.367} & {\color[HTML]{1F1F1F} 0.391} & {\color[HTML]{1F1F1F} 0.321}          & {\color[HTML]{1F1F1F} 0.309}          & {\color[HTML]{1F1F1F} 0.369}          & {\color[HTML]{1F1F1F} 0.427}          & {\color[HTML]{1F1F1F} 0.325}          & {\color[HTML]{1F1F1F} 0.366} & 0.316                 & 0.351                & 0.305    & 0.343             \\
\multirow{-4}{*}{\rotatebox{90}{ETTm2}}         & 720  & 0.511                                  & 0.456                                  & {\color[HTML]{1F1F1F} 0.406}  & {\color[HTML]{1F1F1F} 0.404}           & \cellcolor[HTML]{DDEBF7}\textbf{0.291} & \cellcolor[HTML]{DDEBF7}\textbf{0.374} & {\color[HTML]{1F1F1F} 0.470}           & {\color[HTML]{1F1F1F} 0.439}           & 0.432                 & 0.420                & {\color[HTML]{1F1F1F} 0.502} & {\color[HTML]{1F1F1F} 0.491}          & {\color[HTML]{1F1F1F} 0.420}          & {\color[HTML]{1F1F1F} 0.410}          & {\color[HTML]{1F1F1F} 0.451} & {\color[HTML]{1F1F1F} 0.432} & {\color[HTML]{1F1F1F} 0.408}          & {\color[HTML]{1F1F1F} 0.403}          & {\color[HTML]{1F1F1F} 0.554}          & {\color[HTML]{1F1F1F} 0.522}          & {\color[HTML]{1F1F1F} 0.421}          & {\color[HTML]{1F1F1F} 0.415} & 0.414                 & 0.407                & 0.410    & 0.405             \\
 \midrule
                                & 96   & 0.205                                  & 0.300                                  & 0.207                         & 0.303                                  & 0.187                                  & 0.282                                  & \cellcolor[HTML]{DDEBF7}0.156          & \cellcolor[HTML]{DDEBF7}0.247          & 0.421                 & 0.504                & {\color[HTML]{1F1F1F} 0.181} & {\color[HTML]{1F1F1F} 0.265}          & {\color[HTML]{1F1F1F} 0.196}          & {\color[HTML]{1F1F1F} 0.287}          & {\color[HTML]{1F1F1F} 0.198} & {\color[HTML]{1F1F1F} 0.290} & {\color[HTML]{1F1F1F} 0.168}          & {\color[HTML]{1F1F1F} 0.272}          & {\color[HTML]{1F1F1F} 0.197}          & {\color[HTML]{1F1F1F} 0.282}          & {\color[HTML]{1F1F1F} 0.193}          & {\color[HTML]{1F1F1F} 0.308} & \textbf{0.148}        & \textbf{0.240}       & 0.186    & 0.270             \\
                                & 192  & 0.213                                  & 0.305                                  & 0.215                         & 0.306                                  & 0.191                                  & 0.281                                  & \cellcolor[HTML]{DDEBF7}0.176          & \cellcolor[HTML]{DDEBF7}0.266          & 0.423                 & 0.499                & {\color[HTML]{1F1F1F} 0.184} & {\color[HTML]{1F1F1F} 0.269}          & {\color[HTML]{1F1F1F} 0.199}          & {\color[HTML]{1F1F1F} 0.291}          & {\color[HTML]{1F1F1F} 0.202} & {\color[HTML]{1F1F1F} 0.293} & {\color[HTML]{1F1F1F} 0.184}          & {\color[HTML]{1F1F1F} 0.289}          & {\color[HTML]{1F1F1F} 0.196}          & {\color[HTML]{1F1F1F} 0.285}          & {\color[HTML]{1F1F1F} 0.201}          & {\color[HTML]{1F1F1F} 0.315} & \textbf{0.166}        & \textbf{0.258}       & 0.190    & 0.274             \\
                                & 336  & 0.253                                  & 0.345                                  & 0.225                         & 0.316                                  & 0.210                                 & 0.300                                  & \cellcolor[HTML]{DDEBF7}\textbf{0.193} & \cellcolor[HTML]{DDEBF7}\textbf{0.285} & 0.451                 & 0.528                & {\color[HTML]{1F1F1F} 0.200} & {\color[HTML]{1F1F1F} 0.285}          & {\color[HTML]{1F1F1F} 0.214}          & {\color[HTML]{1F1F1F} 0.305}          & {\color[HTML]{1F1F1F} 0.223} & {\color[HTML]{1F1F1F} 0.318} & {\color[HTML]{1F1F1F} 0.198}          & {\color[HTML]{1F1F1F} 0.300}          & {\color[HTML]{1F1F1F} 0.209}          & {\color[HTML]{1F1F1F} 0.301}          & {\color[HTML]{1F1F1F} 0.214}          & {\color[HTML]{1F1F1F} 0.329} & 0.179                 & 0.272                & 0.206    & 0.293             \\
\multirow{-4}{*}{\rotatebox{90}{Electricity}}   & 720  & 0.250                                  & 0.335                                  & 0.264                         & 0.344                                  & 0.252                                  & 0.334                                  & \cellcolor[HTML]{DDEBF7}0.221          & \cellcolor[HTML]{DDEBF7}\textbf{0.310} & 0.494                 & 0.550                & {\color[HTML]{1F1F1F} 0.236} & 0.318 & {\color[HTML]{1F1F1F} 0.254}          & {\color[HTML]{1F1F1F} 0.335}          & {\color[HTML]{1F1F1F} 0.259} & {\color[HTML]{1F1F1F} 0.341} & {\color[HTML]{1F1F1F} \textbf{0.220}} & {\color[HTML]{1F1F1F} 0.320}          & {\color[HTML]{1F1F1F} 0.245}          & {\color[HTML]{1F1F1F} 0.333}          & {\color[HTML]{1F1F1F} 0.246}          & {\color[HTML]{1F1F1F} 0.355} & 0.209                 & 0.298                & 0.247    & 0.324             \\
 \midrule
                                & 96   & \cellcolor[HTML]{DDEBF7}\textbf{0.163} & \cellcolor[HTML]{DDEBF7}\textbf{0.208} & 0.198                         & 0.238                                  & 0.252                                  & 0.291                                  & 0.199                                  & 0.223                                  & 0.200                 & 0.251                & {\color[HTML]{1F1F1F} 0.175} & {\color[HTML]{1F1F1F} 0.218}          & {\color[HTML]{1F1F1F} 0.177}          & {\color[HTML]{1F1F1F} 0.220}          & {\color[HTML]{1F1F1F} 0.213} & {\color[HTML]{1F1F1F} 0.260} & {\color[HTML]{1F1F1F} 0.172}          & {\color[HTML]{1F1F1F} 0.220}          & {\color[HTML]{1F1F1F} 0.196}          & {\color[HTML]{1F1F1F} 0.255}          & {\color[HTML]{1F1F1F} 0.217}          & {\color[HTML]{1F1F1F} 0.296} & 0.176                 & 0.216                & 0.177    & 0.218             \\
                                & 192  & \cellcolor[HTML]{DDEBF7}\textbf{0.206} & \cellcolor[HTML]{DDEBF7}\textbf{0.249} & 0.242                         & 0.273                                  & 0.300                                  & 0.324                                  & 0.275                                  & 0.279                                  & 0.254                 & 0.294                & {\color[HTML]{1F1F1F} 0.219} & {\color[HTML]{1F1F1F} 0.256}          & {\color[HTML]{1F1F1F} 0.224}          & {\color[HTML]{1F1F1F} 0.260}          & {\color[HTML]{1F1F1F} 0.269} & {\color[HTML]{1F1F1F} 0.300} & {\color[HTML]{1F1F1F} 0.219}          & {\color[HTML]{1F1F1F} 0.261}          & {\color[HTML]{1F1F1F} 0.237}          & {\color[HTML]{1F1F1F} 0.296}          & {\color[HTML]{1F1F1F} 0.276}          & {\color[HTML]{1F1F1F} 0.336} & 0.225                 & 0.257                & 0.225    & 0.259             \\
                                & 336  & \cellcolor[HTML]{DDEBF7}\textbf{0.256} & \cellcolor[HTML]{DDEBF7}\textbf{0.289} & 0.295                         & 0.310                                  & 0.347                                  & 0.353                                  & 0.341                                  & 0.330                                  & 0.311                 & 0.336                & {\color[HTML]{1F1F1F} 0.269} & {\color[HTML]{1F1F1F} 0.296}          & {\color[HTML]{1F1F1F} 0.279}          & {\color[HTML]{1F1F1F} 0.277}          & {\color[HTML]{1F1F1F} 0.330} & {\color[HTML]{1F1F1F} 0.341} & {\color[HTML]{1F1F1F} 0.280}          & {\color[HTML]{1F1F1F} 0.306}          & {\color[HTML]{1F1F1F} 0.283}          & {\color[HTML]{1F1F1F} 0.335}          & {\color[HTML]{1F1F1F} 0.339}          & {\color[HTML]{1F1F1F} 0.380} & 0.281                 & 0.299                & 0.278    & 0.297             \\
\multirow{-4}{*}{\rotatebox{90}{Weather}}       & 720  & \cellcolor[HTML]{DDEBF7}\textbf{0.327} & \cellcolor[HTML]{DDEBF7}\textbf{0.336} & 0.370                         & 0.358                                  & 0.416                                  & 0.395                                  & 0.452                                  & 0.397                                  & 0.379                 & 0.375                & {\color[HTML]{1F1F1F} 0.337} & {\color[HTML]{1F1F1F} 0.344}          & {\color[HTML]{1F1F1F} 0.354}          & {\color[HTML]{1F1F1F} 0.347}          & {\color[HTML]{1F1F1F} 0.404} & {\color[HTML]{1F1F1F} 0.389} & {\color[HTML]{1F1F1F} 0.365}          & {\color[HTML]{1F1F1F} 0.359}          & {\color[HTML]{1F1F1F} 0.345}          & {\color[HTML]{1F1F1F} 0.381}          & {\color[HTML]{1F1F1F} 0.403}          & {\color[HTML]{1F1F1F} 0.428} & 0.358                 & 0.350                & 0.354    & 0.348             \\
 \midrule
                                & 96   & \cellcolor[HTML]{DDEBF7}\textbf{0.085}          & 0.223                                  & 0.094                         & \cellcolor[HTML]{DDEBF7}\textbf{0.203} & 0.150                                  & 0.281                                  & 0.156                                  & 0.247                                  & 0.101                 & 0.223                & {\color[HTML]{1F1F1F} 0.118} & {\color[HTML]{1F1F1F} 0.265}          & {\color[HTML]{1F1F1F} 0.096}          & {\color[HTML]{1F1F1F} 0.219}          & {\color[HTML]{1F1F1F} 0.137} & {\color[HTML]{1F1F1F} 0.260} & {\color[HTML]{1F1F1F} 0.107}          & {\color[HTML]{1F1F1F} 0.234}          & {\color[HTML]{1F1F1F} 0.088} & {\color[HTML]{1F1F1F} 0.218}          & {\color[HTML]{1F1F1F} 0.148}          & {\color[HTML]{1F1F1F} 0.278} & 0.086                 & 0.206                & 0.109    & 0.236             \\  
                                & 192  & \cellcolor[HTML]{DDEBF7}0.190          & 0.336                                 & 0.194                         & \cellcolor[HTML]{DDEBF7}\textbf{0.304} & 0.247                                  & 0.362                                  & 0.298                                  & 0.388                                  & 0.193                 & 0.311                & {\color[HTML]{1F1F1F} 0.179} & {\color[HTML]{1F1F1F} 0.324}          & {\color[HTML]{1F1F1F} \textbf{0.187}} & {\color[HTML]{1F1F1F} 0.309}          & {\color[HTML]{1F1F1F} 0.222} & {\color[HTML]{1F1F1F} 0.341} & {\color[HTML]{1F1F1F} 0.226}          & {\color[HTML]{1F1F1F} 0.344}          & {\color[HTML]{1F1F1F} 0.176}          & {\color[HTML]{1F1F1F} 0.315}          & {\color[HTML]{1F1F1F} 0.271}          & {\color[HTML]{1F1F1F} 0.380} & 0.181                 & 0.304                & 0.205    & 0.327             \\
                                & 336  & 0.484                                  & 0.549                                  & \cellcolor[HTML]{DDEBF7}0.341 & \cellcolor[HTML]{DDEBF7}0.421          & 0.390                                  & 0.460                                  & 0.579                                  & 0.542                                  & 0.358                 & 0.435                & {\color[HTML]{1F1F1F} 0.404} & {\color[HTML]{1F1F1F} 0.506}          & {\color[HTML]{1F1F1F} \textbf{0.327}} & {\color[HTML]{1F1F1F} \textbf{0.415}} & {\color[HTML]{1F1F1F} 0.372} & {\color[HTML]{1F1F1F} 0.447} & {\color[HTML]{1F1F1F} 0.367}          & {\color[HTML]{1F1F1F} 0.448}          & {\color[HTML]{1F1F1F} 0.313}          & {\color[HTML]{1F1F1F} 0.427}          & {\color[HTML]{1F1F1F} 0.460}          & {\color[HTML]{1F1F1F} 0.500} & 0.338                 & 0.422                & 0.356    & 0.436             \\
\multirow{-4}{*}{\rotatebox{90}{Exchange}}      & 720  & \cellcolor[HTML]{DDEBF7}\textbf{0.776} & 0.732                                  & 0.891                         & \cellcolor[HTML]{DDEBF7}0.714          & 0.939                                  & 0.739                                  & 1.161                                  & 0.799                                  & 0.941                 & 0.721                & 0.959                        & 0.805                                 & {\color[HTML]{1F1F1F} 0.875}          & {\color[HTML]{1F1F1F} 0.701}          & {\color[HTML]{1F1F1F} 0.912} & {\color[HTML]{1F1F1F} 0.727} & {\color[HTML]{1F1F1F} 0.964}          & {\color[HTML]{1F1F1F} 0.746}          & {\color[HTML]{1F1F1F} 0.839}          & {\color[HTML]{1F1F1F} \textbf{0.695}} & {\color[HTML]{1F1F1F} 1.195}          & {\color[HTML]{1F1F1F} 0.841} & 0.853                 & 0.696                & 0.901    & 0.716             \\
\midrule
\multicolumn{2}{c|}{$1^{st}$ Count}          & \multicolumn{2}{c|}{\colorbox[HTML]{DDEBF7}{\textbf{19}}}                                                 & \multicolumn{2}{c}{{\color[HTML]{1F1F1F} 4}}                           & \multicolumn{2}{c|}{{\color[HTML]{1F1F1F} 11}}                                   & \multicolumn{2}{c|}{3}                                                           & \multicolumn{2}{c|}{{\color[HTML]{1F1F1F} 0}} & \multicolumn{2}{c|}{{\color[HTML]{1F1F1F} 0}}                         & \multicolumn{2}{c|}{{\color[HTML]{1F1F1F} 3}}                                  & \multicolumn{2}{c|}{{\color[HTML]{1F1F1F} 0}}                & \multicolumn{2}{c|}{{\color[HTML]{1F1F1F} 3}}                                  & \multicolumn{2}{c|}{{\color[HTML]{1F1F1F} 2}}                                  & \multicolumn{2}{c|}{{\color[HTML]{1F1F1F} 3}}                         & \multicolumn{2}{c|}{{\color[HTML]{1F1F1F} 4}} & \multicolumn{2}{c}{4}        \\
\midrule
\multicolumn{2}{c|}{$1^{st}$ Count in FL-FM} & \multicolumn{2}{c|}{\colorbox[HTML]{DDEBF7}{\textbf{28}}}                                                 & \multicolumn{2}{c|}{9}                                                  & \multicolumn{2}{c|}{11}                                                          & \multicolumn{2}{c|}{8}                                                           & \multicolumn{2}{c|}{0}                        & \multicolumn{2}{c|}{-}                                                & \multicolumn{2}{c|}{-}                                                         & \multicolumn{2}{c|}{-}                                       & \multicolumn{2}{c|}{-}                                                         & \multicolumn{2}{c|}{-}                                                         & \multicolumn{2}{c|}{-}                                                & \multicolumn{2}{c|}{-}                        & \multicolumn{2}{c}{-}    \\
\midrule
\multicolumn{2}{c|}{Comm. Params.} & \multicolumn{2}{c|}{\textbf{\colorbox[HTML]{DDEBF7}{0.016 M}}}                                                 & \multicolumn{2}{c|}{6.811 M}                                                  & \multicolumn{2}{c|}{0.538 M}                                                          & \multicolumn{2}{c|}{9.557 M}                                                           & \multicolumn{2}{c|}{0.549 M}                        & \multicolumn{2}{c|}{-}                                                & \multicolumn{2}{c|}{-}                                                         & \multicolumn{2}{c|}{-}                                       & \multicolumn{2}{c|}{-}                                                         & \multicolumn{2}{c|}{-}                                                         & \multicolumn{2}{c|}{-}                                                & \multicolumn{2}{c|}{-}                        & \multicolumn{2}{c}{-}    \\

\bottomrule
\end{tabular}%
}
\end{table*}

\begin{table}[th]
\tiny
\centering
\tabcolsep=0.08cm
\renewcommand\arraystretch{0.8}
\caption{Few-shot forecasting performance. Comparison results under forecasting horizons $F_i\in\{96,192,336,720\}$. Results are averaged over the four prediction lengths. \textbf{Bold} indicates the best performance among all methods. Complete results are reported in Table~\ref{tab:few_shot_full_result}.}
\label{tab:few_shot_result}
\resizebox{1\columnwidth}{!}{%
\begin{tabular}{c|c|c|c|c|c|c|c|c}
\toprule
\rowcolor{blue!15}\multicolumn{9}{c}{\textbf{Few-shot   Long-term Forecasting (5\%)}}
\\
\midrule
Type                    & Method                           & Metric & ETTm1          & ETTm2          & Electricity    & Weather        & Exchange       & $1^{st}$ Count                   \\
\midrule
\multirow{10}{*}{FL-FM} & \multirow{2}{*}{\textsc{FeDPM}}            & MSE    & \textbf{0.538} & 0.310          & 0.248          & \textbf{0.257} & \textbf{0.155} & \multirow{2}{*}{\textbf{6}} \\
                        &                                  & MAE    & \textbf{0.480} & 0.338          & 0.337          & \textbf{0.290} & \textbf{0.293} &                             \\
                      
                        & \multirow{2}{*}{Time-FFM}        & MSE    & 0.567          & 0.293          & 0.324          & 0.292          & 0.167          & \multirow{2}{*}{0}          \\
                        &                                  & MAE    & 0.491          & 0.333          & 0.403          & 0.318          & 0.289          &                             \\
                     
                        & \multirow{2}{*}{FFTS}            & MSE    & 0.613          & \textbf{0.183} & 0.488          & 0.275          & 0.188          & \multirow{2}{*}{2}          \\
                        &                                  & MAE    & 0.533          & \textbf{0.286} & 0.525          & 0.300          & 0.311          &                             \\
                      
                        & \multirow{2}{*}{FL-iTransformer} & MSE    & 1.080          & 0.465          & \textbf{0.235} & 0.355          & 0.165          & \multirow{2}{*}{2}          \\
                        &                                  & MAE    & 0.674          & 0.430          & \textbf{0.315} & 0.340          & 0.290          &                             \\
                       
                        & \multirow{2}{*}{FL-PatchTST}     & MSE    & 0.900          & 0.329          & 0.258          & 0.301          & 0.180          & \multirow{2}{*}{0}          \\
                        &                                  & MAE    & 0.579          & 0.354          & 0.350          & 0.311          & 0.304          &                             \\
                        \midrule
\multirow{6}{*}{Cen-FM} & \multirow{2}{*}{TOTEM}           & MSE    & 0.905          & 0.633          & 1.030          & 0.304          & 1.619          & \multirow{2}{*}{0}          \\
                        &                                  & MAE    & 0.694          & 0.585          & 0.825          & 0.326          & 1.026          &                             \\
                        & \multirow{2}{*}{UniTime}         & MSE    & 0.714          & 0.314          & 0.298          & 0.288          & 0.442          & \multirow{2}{*}{0}          \\
                        &                                  & MAE    & 0.558          & 0.350          & 0.387          & 0.313          & 0.493          &                             \\
                        & \multirow{2}{*}{Cen-PatchTST}    & MSE    & 0.591          & 0.299          & 0.309          & 0.300          & 0.172          & \multirow{2}{*}{0}          \\
                        &                                  & MAE    & 0.497          & 0.340          & 0.392          & 0.324          & 0.294          &                             \\
                        \midrule
\rowcolor{blue!15}\multicolumn{9}{c}{\textbf{Few-shot Long-term   Forecasting (10\%)}}                                                                                                                            \\
\midrule
\multirow{10}{*}{FL-FM} & \multirow{2}{*}{\textsc{FeDPM}}            & MSE    & \textbf{0.575} & 0.307          & 0.245          & \textbf{0.251} & \textbf{0.185} & \multirow{2}{*}{\textbf{5}} \\
                        &                                  & MAE    & \textbf{0.493} & 0.334          & 0.334          & \textbf{0.280} & 0.319          &                             \\
                        & \multirow{2}{*}{Time-FFM}        & MSE    & 0.593          & 0.294          & 0.266          & 0.288          & 0.230          & \multirow{2}{*}{0}          \\
                        &                                  & MAE    & 0.500          & 0.335          & 0.343          & 0.314          & 0.337          &                             \\
                        & \multirow{2}{*}{FFTS}            & MSE    & 0.636          & \textbf{0.179} & 0.382          & 0.275          & 0.242          & \multirow{2}{*}{2}          \\
                        &                                  & MAE    & 0.540          & \textbf{0.285} & 0.452          & 0.297          & 0.350          &                             \\
                        & \multirow{2}{*}{FL-iTransformer} & MSE    & 1.180          & 0.373          & \textbf{0.214} & 0.354          & 0.277          & \multirow{2}{*}{2}          \\
                        &                                  & MAE    & 0.689          & 0.378          & \textbf{0.297} & 0.331          & 0.372          &                             \\
                        & \multirow{2}{*}{FL-PatchTST}     & MSE    & 1.220          & 0.304          & 0.252          & 0.274          & 0.204          & \multirow{2}{*}{1}          \\
                        &                                  & MAE    & 0.647          & 0.339          & 0.348          & 0.291          & \textbf{0.312} &                             \\
                        \midrule
\multirow{6}{*}{Cen-FM} & \multirow{2}{*}{TOTEM}           & MSE    & 0.811          & 0.380          & 0.949          & 0.256          & 0.340          & \multirow{2}{*}{0}          \\
                        &                                  & MAE    & 0.608          & 0.431          & 0.795          & 0.291          & 0.464          &                             \\
                        & \multirow{2}{*}{UniTime}         & MSE    & 0.589          & 0.299          & 0.254          & 0.272          & 0.220          & \multirow{2}{*}{0}          \\
                        &                                  & MAE    & 0.494          & 0.338          & 0.342          & 0.299          & 0.331          &                             \\
                        & \multirow{2}{*}{Cen-PatchTST}    & MSE    & 1.071          & 0.348          & 0.362          & 0.297          & 0.220          & \multirow{2}{*}{0}          \\
                        &                                  & MAE    & 0.662          & 0.378          & 0.429          & 0.316          & 0.330          &    \\
                        \bottomrule
\end{tabular}%
}
\end{table}

\subsection{Discussion}

Table \ref{tab:method_comparison} presents a comparison between Time-FFM \cite{time-ffm}, FFTS \cite{FFTS}, and the proposed \textsc{FeDPM}. \textsc{FeDPM} distinguishes itself from existing baselines through three key architectural advantages.

\textbf{(1) Latent Representation.}
A fundamental limitation of existing baselines lies in their latent spaces. Specifically, Time-FFM~\cite{time-ffm} force temporal signals to conform to text-oriented embedding spaces, which can lead to semantic misalignment. FFTS~\cite{FFTS} projects heterogeneous cross-domain time series into unified continuous latent spaces, despite the fact that temporal semantics frequently manifest as discrete and recurring regimes, rendering the model prone to feature space collapse.
% A fundamental limitation of existing baselines lies in their reliance on text-centric or unified continuous latent spaces to model heterogeneous cross-domain time series data. Such representations are prone to semantic misalignment and feature space collapse, especially under non-IID and cross-domain settings. 
In contrast, \textsc{FeDPM} introduces discrete prototypical memories, which capture domain-invariant temporal patterns without enforcing continuous mappings across heterogeneous domains.

\textbf{(2) Communication Efficiency.}
The communication overhead of baselines primarily arises from the transmission of large-scale model parameters. By communicating only prototypical memories, \textsc{FeDPM} substantially reduces communication overhead by over \textbf{97.03\%} (Section~\ref{sec:Main_Result}).

\textbf{(3) FM Construction.}
Unlike prior approaches that construct FM through parameter stacking—leading to high model complexity—\textsc{FeDPM} constructs the FM via a unified discrete memory mechanism. As a result, the number of trainable parameters is reduced by over \textbf{20.37\%} compared to existing baselines (Section~\ref{sec:model_analysis}).

\section{Experimental Results}

% We comprehensively compare the proposed \textsc{FeDPM} with state-of-the-art models in FL or centralized settings. The numerical results demonstrate the effectiveness of \textsc{FeDPM} in time series forecasting.

\paragraph{Baselines.}
We compare our method against a comprehensive set of representative baselines, covering three categories:
\textbf{(1) Federated Learning of Time Series Foundation Models (FL-FM).}  
These methods are designed specifically for the federated learning setting, including Time-FFM  \cite{time-ffm}, FFTS \cite{FFTS}, FL-iTransformer, and FL-PatchTST.
\textbf{(2) Centralized Time Series Foundation Models (Cen-FM).} 
This category includes foundation models trained under centralized settings, such as TOTEM \cite{totem}, UniTime \cite{unitime}, and Cen-PatchTST. 
\textbf{(3) Centralized Expert Models (Expert).}
These are dataset-specific forecasting models trained from scratch in a centralized manner, including TimesNet \cite{timesnet}, DLinear \cite{dlinear}, FEDformer \cite{fedformer}, iTransformer \cite{itransformer}, and PatchTST  \cite{patchtst}. All baseline models are implemented using the optimal hyperparameters reported in their original papers.
Further details on FL-iTransformer, FL-PatchTST, Cen-PatchTST, and FFTS are provided in Appendix~\ref{sec:Experimental_Details}.

\paragraph{Setup.}
We evaluate on 7 benchmark datasets from various domains: ETTh1, ETTh2, ETTm1, ETTm2, Electricity, Weather, and Exchange, which have been widely adopted for time series forecasting \cite{time-ffm,time-vlm}. Each dataset corresponds to a FL client. 
Detailed introduction of implementation and datasets can be found in Appendix \ref{sec:Experimental_Details}. We use Mean Square Error (MSE) and Mean Absolute Error (MAE) as the evaluation metrics. 
For all domains, the patch length and stride are fixed to $S_n = 4$. 
% The shared cluster limit is set to $\gamma = 0.95$, and the similarity threshold is $\delta = 0.7$. 
The prototypical memory is configured with size $M = 256$ and embedding dimension $D = 64$. Additional hyperparameter settings are reported in Appendix \ref{sec:Experimental_Details}.

\subsection{Main Results} \label{sec:Main_Result}
The main forecasting results are reported in Table~\ref{tab:Full-shot Result}. \textsc{FeDPM} achieves the highest number of first-place rankings among all compared methods, including it in the FL-FM category. Compared with the strongest baseline FFTS, \textsc{FeDPM} reduces MAE by an average of 4.92\%.
More importantly, \textsc{FeDPM} achieves a significantly lower communication cost, requiring \textbf{97.03\% fewer} transmitted parameters than the baseline with the minimal communication overhead.
This efficiency stems from transmitting only local prototypical memories, rather than full model parameters as in existing FL approaches. Since communication overhead is widely recognized as the primary bottleneck in FL systems \cite{communication}, the proposed prototypical memory transfer mechanism offers a more scalable and communication-efficient solution for federated time series forecasting.
These results validate the effectiveness of the proposed prototypical memory transfer framework, which enables the identification and exploitation of domain-relevant knowledge for improved forecasting performance.

\subsection{Few-Shot Forecasting}
In this part, we evaluate the few-shot forecasting capability of \textsc{FeDPM}, and results are reported in Table~\ref{tab:few_shot_result}. Specifically, we compare its performance with FL-FM and Cen-FM baselines under few-shot settings, where only 5\% and 10\% of the data are used for training, following the protocols in \cite{GPT4TS,time-llm,time-vlm,time-ffm}.
Under the 5\% training setting, \textsc{FeDPM} achieves a 7.29\% MAE reduction compared with the strongest baseline FFTS, while under the 10\% setting, it also reduces MAE by 6.42\%. These results demonstrate that \textsc{FeDPM} maintains strong forecasting performance even with limited training data, highlighting the effectiveness of the proposed prototypical memory transfer mechanism, which enables the model to leverage transferable temporal patterns from other domains to improve predictions.

\subsection{Model Analysis}\label{sec:model_analysis}

\paragraph{Model Ablation.}
\begin{table}[tbh!]

\tiny
\centering
\tabcolsep=0.08cm
\renewcommand\arraystretch{0.8}

% \centering
% \setlength{\tabcolsep}{1.5pt}  % 减少列间距(默认6pt)
% \renewcommand{\arraystretch}{1.1}  % 压缩行高
\caption{Ablation results on seven datasets with forecasting horizons $F_i \in \{96, 192\}$. 
\label{Tab:Ablation_Result}
All results are averaged over the two prediction lengths. \textbf{Bold} denotes the best performance.}
% \resizebox{0.85\columnwidth}{!}{%
\begin{tabular*}{\columnwidth}{@{\extracolsep{\fill}}c|c|c|c|c|c|c|c|c}
% \begin{tabular}{c|c|c|c|c|c|c|c|c}
\toprule
Method                                                                               & Metric & ETTh1          & ETTh2          & ETTm1          & ETTm2          & Electricity    & Weather        & Exchange       \\
\midrule
\multirow{2}{*}{Ours}                                                                & MSE    & \textbf{0.422} & \textbf{3} & \textbf{0.353} & \textbf{0.216} & \textbf{0.209} & \textbf{0.185} & \textbf{0.142} \\
                                                                                     & MAE    & \textbf{0.424} & \textbf{0.368} & \textbf{0.376} & \textbf{0.281} & \textbf{0.303} & \textbf{0.229} & \textbf{0.283} \\
\multirow{2}{*}{\makecell{w/ \\Average}}         & MSE    & 0.441          & 0.350          & 0.359          & 0.231          & 0.232          & 0.218          & 0.177          \\
                                                                                     & MAE    & 0.429          & 0.373          & 0.383          & 0.291          & 0.319          & 0.255          & 0.303          \\
\multirow{2}{*}{\makecell{w/ \\Local Memory}} & MSE    &   0.431             &       0.346         &       0.378         &         0.224       &         0.273       &      0.204          &  0.159              \\
                                                                                     & MAE    &    0.539            &    0.373            & 0.385               &  0.285              &     0.359           &   0.247             &   0.297   \\
    \multirow{2}{*}{\makecell{w/ \\Global Memory}} & MSE    & 0.428          & 0.343          & 0.359          & \textbf{0.216} & 0.224          & 0.186          & \textbf{0.142} \\
                                                                                      & MAE    & 0.428          & 0.369          & 0.384          & 0.283          & 0.316          & 0.230          & \textbf{0.283}\\         
                                                                                     \bottomrule
\end{tabular*}%
% }
\end{table}

\begin{table}[th!]
\centering
\caption{Backbone models of Encoder ablation results on seven datasets with forecasting horizons $F_i \in \{96, 192\}$. All results are averaged over the two prediction lengths. \textbf{Bold} denotes the best performance across all types.}
\label{tab:different_backbone}
\tiny
\centering
\renewcommand\arraystretch{0.8}
\setlength{\tabcolsep}{1pt}  % 减少列间距(默认6pt)
% \resizebox{0.85\columnwidth}{!}{%
\begin{tabular*}{\columnwidth}{@{\extracolsep{\fill}}c|c|c|c|c|c|c|c|c|c}
\toprule
Type                      & Method                    & Metric & \multicolumn{1}{c|}{ETTh1} & \multicolumn{1}{c|}{ETTh2} & \multicolumn{1}{c|}{ETTm1} & \multicolumn{1}{c|}{ETTm2} & \multicolumn{1}{c|}{Electricity} & \multicolumn{1}{c|}{Weather} & \multicolumn{1}{c}{Exchange} \\
\midrule
\multirow{8}{*}{\rotatebox{90}{\textsc{FeDPM} Variants}}     & \multirow{2}{*}{Transformer}     & MSE    & 0.422                     & 0.342                     & \textbf{0.353}            & 0.216                     & \textbf{0.209}                  & \textbf{0.185}              & 0.142                        \\
                          &                           & MAE    & \textbf{0.424}            & \textbf{0.368}            & \textbf{0.376}            & \textbf{0.281}            & \textbf{0.303}                  & \textbf{0.229}              & 0.283                        \\
                          & \multirow{2}{*}{CNN}      & MSE    & 0.427                     & 0.344                     & 0.363                     & 0.219                     & 0.220                           & 0.187                       & 0.144                        \\
                          &                           & MAE    & 0.435                     & 0.373                     & 0.387                     & 0.287                     & 0.316                           & 0.231                       & 0.282                        \\
                          & \multirow{2}{*}{FC}       & MSE    & 0.700                     & 0.360                     & 0.690                     & 0.235                     & 0.842                           & 0.200                       & 0.146                        \\
                          &                           & MAE    & 0.563                     & 0.390                     & 0.553                     & 0.313                     & 0.754                           & 0.251                       & 0.284                        \\
                          & \multirow{2}{*}{RNN}      & MSE    & \textbf{0.421}            & \textbf{0.339}            & 0.361                     & \textbf{0.214}            & 0.221                           & 0.186                       & \textbf{0.139}               \\
                          &                           & MAE    & 0.428                     & 0.369                     & 0.387                     & \textbf{0.281}            & 0.315                           & 0.231                       & \textbf{0.278}               \\
                          \midrule
\multirow{4}{*}{\rotatebox{90}{Baseline}} & \multirow{2}{*}{Time-FFM} & MSE    & 0.433                     & 0.343                     & 0.378                     & \textbf{0.214}            & 0.211                           & 0.220                       & 0.144                        \\
                          &                           & MAE    & 0.426                     & 0.374                     & 0.383                     & 0.289                     & 0.305                           & 0.256                       & 0.254                        \\
                          & \multirow{2}{*}{TOTEM}    & MSE    & 0.430                     & 0.344                     & 0.393                     & 0.227                     & 0.183                           & 0.197                       & 0.149                        \\
                          &                           & MAE    & 0.421                     & 0.369                     & 0.397                     & 0.294                     & 0.267                           & 0.237                       & 0.294   \\
                          \bottomrule
\end{tabular*}%
\end{table}
We conduct extensive ablation studies on the proposed \textsc{FeDPM} framework, and the results are summarized in Table~\ref{Tab:Ablation_Result}. 
First, we replace the proposed \textit{Cross-Domain Memory Update} Module with the average method (denoted as \textit{w/ Average}) to evaluate the effectiveness of semantic-aware aggregation. 
The results show that substituting our aggregation strategy with Average strategy leads to an average performance degradation of 7.18\%, even when the transmitted memories preserve their original ordering. 
If the memories ordering is further disrupted, the prediction accuracy degrades even more severely.

In addition, we consider a variant where local memories are not uploaded to the server and are kept entirely local (\textit{w/ Local Memory}) to assess the contribution of cross-domain knowledge sharing. 
Under this setting, the average prediction performance drops by 9.34\%, indicating that the cross-domain prototypical knowledge can complement each other.
This observation suggests that leveraging complementary patterns from other domains effectively enhances forecasting accuracy. 

We further evaluate a variant where all domains rely solely on the global memory without personalized memory components (\textit{w/ Global Memory}). This variant results in an average performance drop of 1.43\%, which is consistent with our analysis that each domain contains both shareable and domain-specific knowledge.

\paragraph{Encoder Ablation.}
We evaluate \textsc{FeDPM} using different encoder backbone architectures \cite{gru,RAL-GRU,cnn1d}.
As shown in Table \ref{tab:different_backbone}, \textsc{FeDPM} achieves superior performance over the baseline in the majority of cases across diverse encoder backbones, highlighting the robustness and general applicability of the proposed framework. Given that the Transformer encoder yields the best overall performance, we adopt it as the default encoder backbone in all subsequent experiments.

\paragraph{Model Efficiency.}
\begin{figure}[t]
    \centering
    \includegraphics[width=1.0\columnwidth]{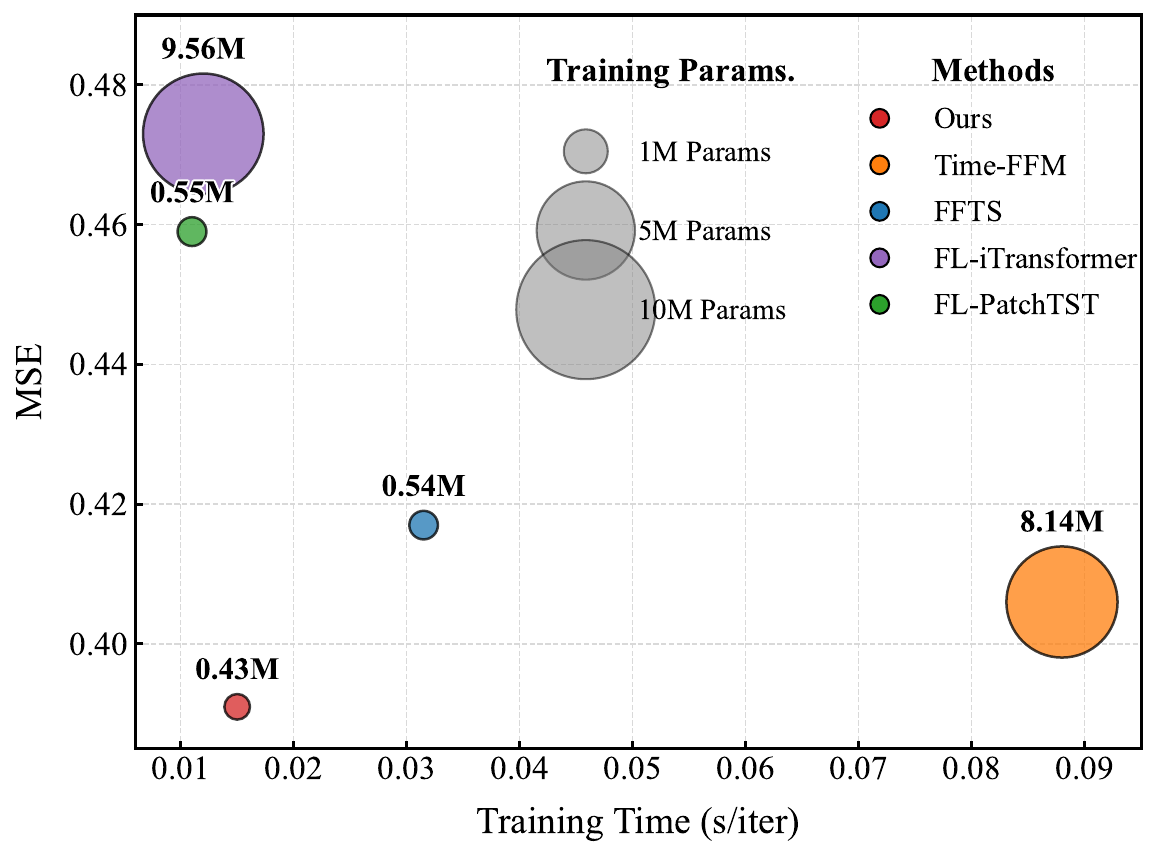}
    \caption{Model efficiency comparison on ETTh1 ($F_i$=96) in terms of forecasting MSE, training time, and training Parameters.}
    \label{fig:Efficiency}
\end{figure}
% \begin{table}[h!]
%   \centering
%   \caption{Efficiency analysis of \textsc{FeDPM} on ETTh1 dataset. \textbf{Bold}: the best.}
%   \label{tab:efficiency_comparison}
%   \resizebox{1\columnwidth}{!}{%
%   \begin{tabular}{cccc}
%     \toprule
%     % 第一行表头
%     \multirow{2}{*}{\textbf{Method}} & \textbf{Training Param.} & \textbf{Training Time} & \textbf{Comm. Param.} \\
%      & \textbf{(M)} & \textbf{(s/iter)} & \textbf{(M)} \\
%     \midrule
%     \textsc{FeDPM} & \textbf{0.432} & 0.015 & \textbf{0.016} \\
%     Time-FFM & 8.138 & 0.088 & 6.811 \\
%     FFTS & 0.538 &  0.032& 0.538\\
%     FL-iTransformer & 9.557 & 0.012 & 9.557 \\
%     FL-PatchTST & 0.549 & \textbf{0.011} & 0.549 \\
%     \bottomrule
%   \end{tabular}
%   }
% \end{table}

Figure \ref{fig:Efficiency} demonstrates that \textsc{FeDPM} achieves state-of-the-art performance with the fewest trainable parameters among all compared methods, yielding a parameter reduction of over 20.37\%. In addition, \textsc{FeDPM} exhibits substantially lower training time than Time-FFM and FFTS, while remaining comparable to other federated baselines, including FL-iTransformer and FL-PatchTST.

\paragraph{Privacy Preservation.}

\begin{table}[!t]
\centering
\setlength{\tabcolsep}{2pt}  % 减少列间距(默认6pt)
\renewcommand{\arraystretch}{1}  % 压缩行高
\caption{Performance of \textsc{FeDPM} under different privacy-preserving mechanisms across forecasting horizons $F_i\in\{96,192,336,720\}$. 
\colorbox[HTML]{D9E1F2}{Blue}: best result among \textsc{FeDPM} variants with noise injection; 
\colorbox[HTML]{E2EFDA}{Green}: best result among baseline methods; 
\textbf{Bold}: best result across all methods.}

\label{tab:Noise_Sensitity}
\resizebox{1.0\columnwidth}{!}{%
\begin{tabular}{cc|cc|cc|cc|cc|cc|cc|cc}
\toprule
\multicolumn{2}{c|}{Type}        & \multicolumn{6}{c|}{\textsc{FeDPM}$+$ Noise}                                                                                                                                                                    & \multicolumn{8}{c}{Baseline (w/o Noise)}                                                                                                                                                                                                                                                                 \\
\midrule
\multicolumn{2}{c|}{Method}      & \multicolumn{2}{c|}{Gaussian}                                           & \multicolumn{2}{c|}{Exponential}                & \multicolumn{2}{c|}{Laplace}                                            & \multicolumn{2}{c|}{FL-iTransformer}                          & \multicolumn{2}{c|}{FL-PatchTST} & \multicolumn{2}{c|}{UniTime}                                                                                                   & \multicolumn{2}{c}{Cen-PatchTST}                            \\
\midrule
\multicolumn{2}{c|}{Metric}      & MSE                           & MAE                                    & MSE                                    & MAE   & MSE                                    & MAE                           & MSE                          & MAE                           & MSE            & MAE            & MSE                                                           & MAE                                                           & MSE                          & MAE                          \\
\midrule
                          & 96  & 0.180                         & \cellcolor[HTML]{D9E1F2}0.231          & \cellcolor[HTML]{D9E1F2}0.179          & 0.237 & 0.199                                  & 0.249                         & 0.199                        & \cellcolor[HTML]{E2EFDA}0.223 & 0.200          & 0.251          & \cellcolor[HTML]{E2EFDA}{\color[HTML]{1F1F1F} \textbf{0.177}} & \cellcolor[HTML]{E2EFDA}{\color[HTML]{1F1F1F} \textbf{0.220}} & {\color[HTML]{1F1F1F} 0.213} & {\color[HTML]{1F1F1F} 0.260} \\
                          & 192 & 0.250                         & 0.293                                  & 0.221                                  & 0.275 & \cellcolor[HTML]{D9E1F2}\textbf{0.219} & \cellcolor[HTML]{D9E1F2}0.264 & 0.275                        & 0.279                         & 0.254          & 0.294          & \cellcolor[HTML]{E2EFDA}{\color[HTML]{1F1F1F} 0.224}          & \cellcolor[HTML]{E2EFDA}{\color[HTML]{1F1F1F} \textbf{0.260}} & {\color[HTML]{1F1F1F} 0.269} & {\color[HTML]{1F1F1F} 0.300} \\
                          & 336 & 0.288                         & 0.321                                  & \cellcolor[HTML]{D9E1F2}\textbf{0.271} & 0.312 & 0.272                                  & \cellcolor[HTML]{D9E1F2}0.307 & 0.341                        & 0.330                         & 0.311          & 0.336          & \cellcolor[HTML]{E2EFDA}{\color[HTML]{1F1F1F} 0.279}          & \cellcolor[HTML]{E2EFDA}{\color[HTML]{1F1F1F} \textbf{0.277}} & {\color[HTML]{1F1F1F} 0.330} & {\color[HTML]{1F1F1F} 0.341} \\
\multirow{-4}{*}{\rotatebox{90}{Weather}} & 720 & 0.345                         & \cellcolor[HTML]{D9E1F2}0.355          & \cellcolor[HTML]{D9E1F2}\textbf{0.340} & 0.359 & 0.363                                  & 0.374                         & 0.452                        & 0.397                         & 0.379          & 0.375          & \cellcolor[HTML]{E2EFDA}{\color[HTML]{1F1F1F} \textbf{0.354}} & \cellcolor[HTML]{E2EFDA}{\color[HTML]{1F1F1F} 0.347}          & {\color[HTML]{1F1F1F} 0.404} & {\color[HTML]{1F1F1F} 0.389} \\
\midrule
                          & 96  & \cellcolor[HTML]{D9E1F2}0.184 & \cellcolor[HTML]{D9E1F2}\textbf{0.263} & 0.202                                  & 0.292 & 0.204                                  & 0.286                         & 0.212                        & 0.277                         & 0.195          & 0.282          & \cellcolor[HTML]{E2EFDA}{\color[HTML]{1F1F1F} \textbf{0.183}} & \cellcolor[HTML]{E2EFDA}{\color[HTML]{1F1F1F} 0.266}          & {\color[HTML]{1F1F1F} 0.240} & {\color[HTML]{1F1F1F} 0.318} \\
                          & 192 & \cellcolor[HTML]{D9E1F2}0.254 & \cellcolor[HTML]{D9E1F2}0.311          & 0.275                                  & 0.338 & 0.255                                  & 0.311                         & 0.282                        & 0.325                         & 0.262          & 0.318          & \cellcolor[HTML]{E2EFDA}{\color[HTML]{1F1F1F} \textbf{0.251}} & \cellcolor[HTML]{E2EFDA}{\color[HTML]{1F1F1F} \textbf{0.310}} & {\color[HTML]{1F1F1F} 0.301} & {\color[HTML]{1F1F1F} 0.352} \\
                          & 336 & \cellcolor[HTML]{D9E1F2}0.325 & \cellcolor[HTML]{D9E1F2}0.357          & 0.346                                  & 0.380 & \cellcolor[HTML]{D9E1F2}0.325          & 0.358                         & 0.351                        & 0.372                         & 0.320          & 0.353          & \cellcolor[HTML]{E2EFDA}{\color[HTML]{1F1F1F} \textbf{0.319}} & \cellcolor[HTML]{E2EFDA}{\color[HTML]{1F1F1F} \textbf{0.351}} & {\color[HTML]{1F1F1F} 0.367} & {\color[HTML]{1F1F1F} 0.391} \\
\multirow{-4}{*}{\rotatebox{90}{ETTm2}}   & 720 & 0.447                         & \cellcolor[HTML]{D9E1F2}0.426          & \cellcolor[HTML]{D9E1F2}0.436          & 0.448 & 0.478                                  & 0.456                         & {\color[HTML]{1F1F1F} 0.470} & {\color[HTML]{1F1F1F} 0.439}  & 0.432          & 0.420          & \cellcolor[HTML]{E2EFDA}{\color[HTML]{1F1F1F} \textbf{0.420}} & \cellcolor[HTML]{E2EFDA}{\color[HTML]{1F1F1F} \textbf{0.410}} & {\color[HTML]{1F1F1F} 0.451} & {\color[HTML]{1F1F1F} 0.432} \\
\bottomrule
\end{tabular}%
}
\end{table}

Differential privacy is a widely adopted strategy in federated learning to protect data privacy  \cite{liu2025personalized,FedPA,FedRAP}, and is typically achieved by injecting random noise (e.g., Laplace, Gaussian, or exponential noise) into uploaded model parameters.
In this work, we apply random noise to the communicated local memories in \textsc{FeDPM}. Specifically, we consider Gaussian noise ($\mu=0$, $\lambda=1$), exponential noise ($\lambda=1$), and Laplace noise ($\mu=0$, $\lambda=1$), where $\mu$ and $\lambda$ denote the mean and scale parameters of the corresponding noise distributions, followed \cite{liu2025personalized}. The baseline models are evaluated without noise injection.

Comparison results in Table~\ref{tab:Noise_Sensitity} show that \textsc{FeDPM} remains highly robust under injected noise. Even with noise perturbations, \textsc{FeDPM} achieves performance that is very close to the best results of the baseline methods without noise injection. 
Notably, \textsc{FeDPM} further outperforms the baselines in MSE on the Weather dataset at forecasting horizons of 336 and 720, and in MAE on the ETTm2 dataset at a horizon of 96 and the Weather dataset at a horizon of 192.
% The results show that on the Weather dataset, \textsc{FeDPM} consistently outperforms baseline methods across all metrics and forecasting horizons, even after noise injection. On ETTm2, \textsc{FeDPM} still achieves superior MSE performance at forecasting horizons of 96 and 192. 
These results further demonstrate the robustness of the proposed \textsc{FeDPM} framework under privacy-preserving noise perturbations, indicating its suitability for deployment in privacy-sensitive scenarios while maintaining high predictive accuracy.

\subsection{Case Study}
% \begin{figure}[th]
%     \centering
%     % --- 子图 (a) ---
%     \begin{subfigure}[b]{0.83\linewidth}
%         \centering
%         \includegraphics[width=\linewidth]{Figure/Experiment/intra_client_weather_132-221-227_norm_ICML.pdf}
%         \caption{}
%         \label{fig:vis_temporal}
%     \end{subfigure}
%     \hfill
%     % --- 子图 (b) ---
%     \begin{subfigure}[b]{0.8\linewidth}
%         \centering
%         \includegraphics[width=\linewidth]{Figure/Experiment/tsne_weather_132-221-227.pdf}
%         \caption{}
%         \label{fig:vis_tsne}
%     \end{subfigure}
%     \caption{Visualization of prototypes on the Weather dataset. (a) Input patches ($P_n=4$) associated with representative prototypes in the time domain. (b) Input patches and prototypes projected into the latent space using t-SNE \cite{t-SNE}.}
%     \label{fig:prototype_analysis}
% \end{figure}

\begin{figure}[th]
    \centering
    \includegraphics[width=0.90\columnwidth]{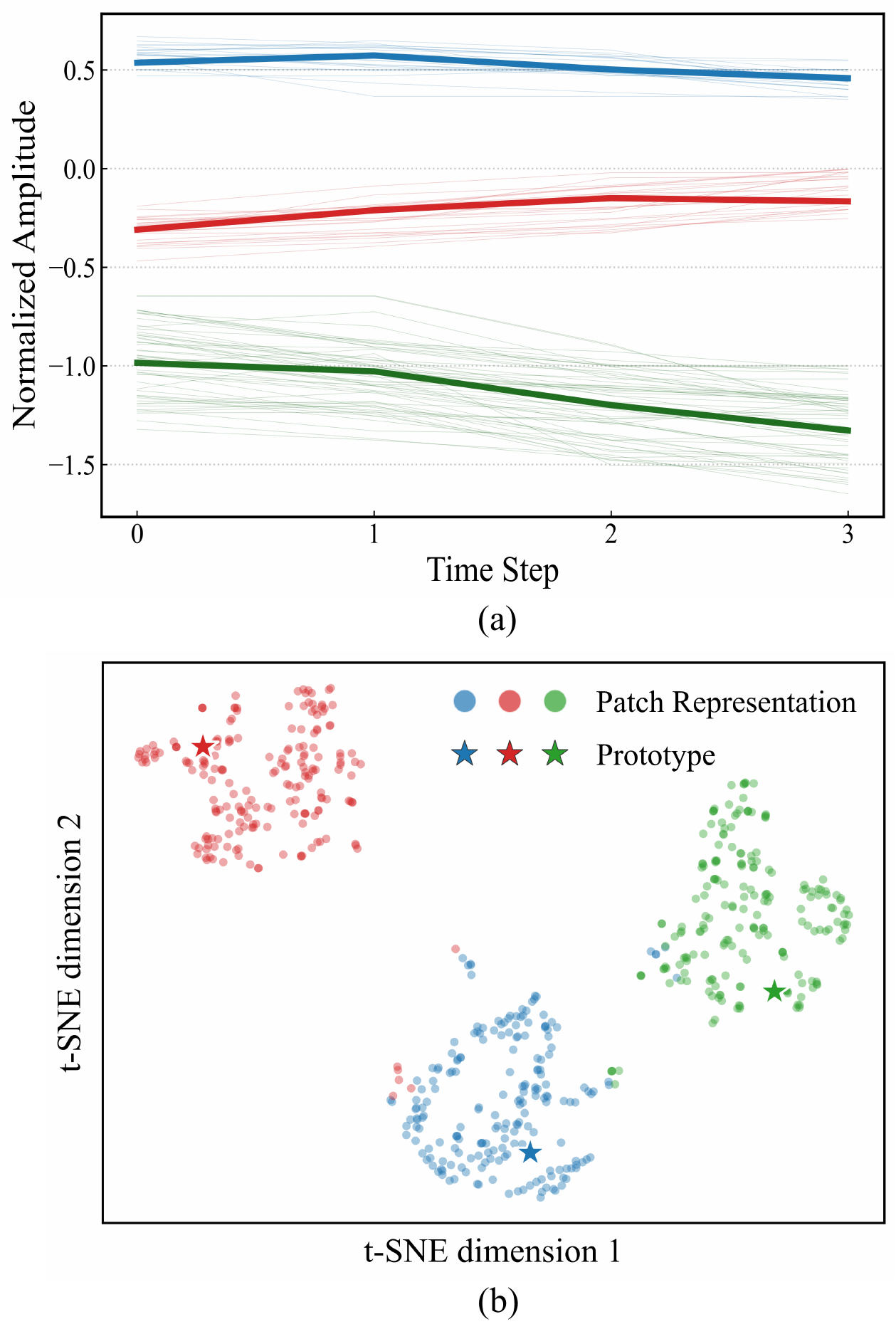}
    \caption{Visualization of prototypes on the Weather dataset. 
    (a) Input patches ($P_n = 4$) and their corresponding representative prototypes in the time domain, where thick lines denote prototypes and thin lines denote input patches.
    (b) Patch Representation and prototype embeddings projected into a shared latent space using t-SNE \cite{t-SNE}.}
    \label{fig:prototype_analysis}
\end{figure}

Figure \ref{fig:prototype_analysis} visualizes input patches from the Weather dataset assigned to three representative prototypes. We employ distinct colors to denote different prototypes: blue, red, and green correspond to prototype 132, 221, and 227, respectively. (a) displays input patches in the original time domain, while (b) projects them into the latent space output by the encoder. Notably, input patches assigned to different prototypes exhibit clearly distinguishable structures in both domains, demonstrating that each prototype effectively captures a unique and disentangled temporal pattern.

\section{Conclusion \& Future Work}
In this work, we identify representation mismatch as a fundamental bottleneck for TSFMs under FL, motivating the need for domain-native and unified discrete representations.
To address this challenge, we propose \textsc{FeDPM}, a parameter- and communication-efficient federated framework that incorporates learnable discrete prototypical memories to balance shared and personalized knowledge. By enabling semantic aggregation across heterogeneous domains without sharing raw data, \textsc{FeDPM} effectively mitigates cross-domain representation misalignment.
Extensive experiments on seven real-world benchmarks show that \textsc{FeDPM} achieves superior performance over existing federated learning baselines, while reducing communication overhead by over 97.03\% and the number of trainable parameters by more than 20.37\%. These results validate both the effectiveness and scalability of \textsc{FeDPM} in practical federated learning scenarios.

\paragraph{Limitations \& Future Works.}
\textsc{FeDPM} has several limitations that warrant further investigation. First, the current framework relies on manual hyperparameter tuning, which limit its adaptability across diverse FL settings. Second, the server-side cross-domain memory alignment module incurs relatively high computational complexity, leading to longer training time and preventing the method from achieving optimal efficiency.
In future work, we will explore adaptive hyperparameter selection mechanisms and more efficient cross-domain memory alignment strategies. In addition, we plan to investigate sparse prototype transmission schemes to further reduce communication costs and improve scalability.

\nocite{langley00}

\newpage
\section*{Impact Statement}
This work aims to advance the field of machine learning by supporting collaborative time series forecasting in privacy-sensitive domains, such as healthcare (e.g., disease transmission modeling) and critical infrastructure (e.g., energy grid management), without requiring the exchange of raw data. By enabling cross-domain knowledge sharing while limiting direct data exposure, the proposed approach may help mitigate privacy risks commonly associated with centralized data collection.

Empirical results suggest that the method remains robust under standard privacy-preserving noise mechanisms. We do not anticipate immediate negative societal impacts arising from this work; nevertheless, we emphasize the importance of continued research into fairness, robustness, and security when deploying federated learning systems in real-world, high-stakes applications.

\bibliography{reference}

@misc{patchtst,
      title={A Time Series is Worth 64 Words: Long-term Forecasting with Transformers}, 
      author={Yuqi Nie and Nam H. Nguyen and Phanwadee Sinthong and Jayant Kalagnanam},
      year={2023},
      eprint={2211.14730},
      archivePrefix={arXiv},
      primaryClass={cs.LG},
      url={https://arxiv.org/abs/2211.14730}, 
}

@article{san,
  title={Adaptive normalization for non-stationary time series forecasting: A temporal slice perspective},
  author={Liu, Zhiding and Cheng, Mingyue and Li, Zhi and Huang, Zhenya and Liu, Qi and Xie, Yanhu and Chen, Enhong},
  journal={Advances in Neural Information Processing Systems},
  volume={36},
  pages={14273--14292},
  year={2023}
}

@inproceedings{revin,
  title={Reversible instance normalization for accurate time-series forecasting against distribution shift},
  author={Kim, Taesung and Kim, Jinhee and Tae, Yunwon and Park, Cheonbok and Choi, Jang-Ho and Choo, Jaegul},
  booktitle={International conference on learning representations},
  year={2021}
}

@article{transformer,
  title={Attention is all you need},
  author={Vaswani, Ashish and Shazeer, Noam and Parmar, Niki and Uszkoreit, Jakob and Jones, Llion and Gomez, Aidan N and Kaiser, {\L}ukasz and Polosukhin, Illia},
  journal={Advances in neural information processing systems},
  volume={30},
  year={2017}
}

@misc{totem,
      title={TOTEM: TOkenized Time Series EMbeddings for General Time Series Analysis}, 
      author={Sabera Talukder and Yisong Yue and Georgia Gkioxari},
      year={2025},
      eprint={2402.16412},
      archivePrefix={arXiv},
      primaryClass={cs.LG},
      url={https://arxiv.org/abs/2402.16412}, 
}

@article{vit,
  title={An image is worth 16x16 words: Transformers for image recognition at scale},
  author={Dosovitskiy, Alexey},
  journal={arXiv preprint arXiv:2010.11929},
  year={2020}
}

@article{gpt3,
  title={Language models are few-shot learners},
  author={Brown, Tom and Mann, Benjamin and Ryder, Nick and Subbiah, Melanie and Kaplan, Jared D and Dhariwal, Prafulla and Neelakantan, Arvind and Shyam, Pranav and Sastry, Girish and Askell, Amanda and others},
  journal={Advances in neural information processing systems},
  volume={33},
  pages={1877--1901},
  year={2020}
}

@article{vqvae,
  title={Neural discrete representation learning},
  author={Van Den Oord, Aaron and Vinyals, Oriol and others},
  journal={Advances in neural information processing systems},
  volume={30},
  year={2017}
}

@inproceedings{smoothl1,
  title={Fast r-cnn},
  author={Girshick, Ross},
  booktitle={Proceedings of the IEEE international conference on computer vision},
  pages={1440--1448},
  year={2015}
}

@inproceedings{fedavg,
  title={Communication-efficient learning of deep networks from decentralized data},
  author={McMahan, Brendan and Moore, Eider and Ramage, Daniel and Hampson, Seth and y Arcas, Blaise Aguera},
  booktitle={Artificial intelligence and statistics},
  pages={1273--1282},
  year={2017},
  organization={PMLR}
}

@article{FedProx,
  title={Federated optimization in heterogeneous networks},
  author={Li, Tian and Sahu, Anit Kumar and Zaheer, Manzil and Sanjabi, Maziar and Talwalkar, Ameet and Smith, Virginia},
  journal={Proceedings of Machine learning and systems},
  volume={2},
  pages={429--450},
  year={2020}
}

@inproceedings{BFS,
  title={A work-efficient parallel breadth-first search algorithm (or how to cope with the nondeterminism of reducers)},
  author={Leiserson, Charles E and Schardl, Tao B},
  booktitle={Proceedings of the twenty-second annual ACM symposium on Parallelism in algorithms and architectures},
  pages={303--314},
  year={2010}
}

@article{cnn1d,
  title={Rethinking 1d-cnn for time series classification: A stronger baseline},
  author={Tang, Wensi and Long, Guodong and Liu, Lu and Zhou, Tianyi and Jiang, Jing and Blumenstein, Michael},
  journal={arXiv preprint arXiv:2002.10061},
  pages={1--7},
  year={2020},
  publisher={Feb}
}

@article{gru,
  title={Empirical evaluation of gated recurrent neural networks on sequence modeling},
  author={Chung, Junyoung and Gulcehre, Caglar and Cho, KyungHyun and Bengio, Yoshua},
  journal={arXiv preprint arXiv:1412.3555},
  year={2014}
}

@inproceedings{attention-based_permutation-invariant,
  title={Set transformer: A framework for attention-based permutation-invariant neural networks},
  author={Lee, Juho and Lee, Yoonho and Kim, Jungtaek and Kosiorek, Adam and Choi, Seungjin and Teh, Yee Whye},
  booktitle={International conference on machine learning},
  pages={3744--3753},
  year={2019},
  organization={PMLR}
}

@inproceedings{timesnet,
  title={Timesnet: Temporal 2d-variation modeling for general time series analysis},
  author={Wu, Haixu and Hu, Tengge and Liu, Yong and Zhou, Hang and Wang, Jianmin and Long, Mingsheng},
  booktitle={The eleventh international conference on learning representations},
  year={2022}
}

@inproceedings{unitime,
  title={UniTime: A Language-Empowered Unified Model for Cross-Domain Time Series Forecasting},
  author={Liu, Xu and Hu, Junfeng and Li, Yuan and Diao, Shizhe and Liang, Yuxuan and Hooi, Bryan and Zimmermann, Roger},
  booktitle={Proceedings of the ACM Web Conference 2024},
  year={2024}
}

@article{llm4ts,
  title={Llm4ts: Two-stage fine-tuning for time-series forecasting with pre-trained llms},
  author={Chang, Ching and Peng, Wen-Chih and Chen, Tien-Fu},
  journal={arXiv preprint arXiv:2308.08469},
  year={2023}
}

@article{time-ffm,
  title={Time-ffm: Towards lm-empowered federated foundation model for time series forecasting},
  author={Liu, Qingxiang and Liu, Xu and Liu, Chenghao and Wen, Qingsong and Liang, Yuxuan},
  journal={Advances in Neural Information Processing Systems},
  volume={37},
  pages={94512--94538},
  year={2024}
}

@article{time-llm,
  title={Time-llm: Time series forecasting by reprogramming large language models},
  author={Jin, Ming and Wang, Shiyu and Ma, Lintao and Chu, Zhixuan and Zhang, James Y and Shi, Xiaoming and Chen, Pin-Yu and Liang, Yuxuan and Li, Yuan-Fang and Pan, Shirui and others},
  journal={arXiv preprint arXiv:2310.01728},
  year={2023}
}

@article{time-vlm,
  title={Time-vlm: Exploring multimodal vision-language models for augmented time series forecasting},
  author={Zhong, Siru and Ruan, Weilin and Jin, Ming and Li, Huan and Wen, Qingsong and Liang, Yuxuan},
  journal={arXiv preprint arXiv:2502.04395},
  year={2025}
}

@inproceedings{FFTS,
  title={Federated foundation models on heterogeneous time series},
  author={Chen, Shengchao and Long, Guodong and Jiang, Jing and Zhang, Chengqi},
  booktitle={Proceedings of the AAAI Conference on Artificial Intelligence},
  volume={39},
  number={15},
  pages={15839--15847},
  year={2025}
}

@article{FedTime,
  title={A federated large language model for long-term time series forecasting},
  author={Abdel-Sater, Raed and Hamza, A Ben},
  journal={arXiv preprint arXiv:2407.20503},
  year={2024}
}

@article{gdpr,
  title={The eu general data protection regulation (gdpr)},
  author={Voigt, Paul and Von dem Bussche, Axel},
  journal={A practical guide, 1st ed., Cham: Springer International Publishing},
  volume={10},
  number={3152676},
  pages={10--5555},
  year={2017},
  publisher={Springer}
}

@article{CCPA,
  title={California consumer privacy act (CCPA)},
  author={Bonta, Rob},
  journal={Retrieved from State of California Department of Justice: https://oag. ca. gov/privacy/ccpa},
  year={2022}
}

@article{time_series_survey,
  title={Deep learning-based time series forecasting},
  author={Song, Xiaobao and Deng, Liwei and Wang, Hao and Zhang, Yaoan and He, Yuxin and Cao, Wenming},
  journal={Artificial Intelligence Review},
  volume={58},
  number={1},
  pages={23},
  year={2024},
  publisher={Springer}
}

@article{RAL-GRU,
  title={Unveiling Uncertainty-Aware Autonomous Cooperative Learning Based Planning Strategy},
  author={Zhang, Shiyao and Deng, Liwei and Zhang, Shuyu and Yuan, Weijie and Zhang, Hong},
  journal={IEEE Robotics and Automation Letters},
  year={2025},
  publisher={IEEE}
}

@ARTICLE{D2Vformer,
    title={D2Vformer: A Flexible Time-Series Prediction Model Based on Time-Position Embedding}, 
    author={Song, Xiaobao and Wang, Hao and Deng, Liwei and Wang, Dong and Qiu, Hongbo and He, Yuxin and Cao, Wenming and Leung, Chi-Sing},
    journal={IEEE Transactions on Neural Networks and Learning Systems}, 
    year={2025},
    publisher={IEEE},
   }

@inproceedings{dlinear,
  title={Are transformers effective for time series forecasting?},
  author={Zeng, Ailing and Chen, Muxi and Zhang, Lei and Xu, Qiang},
  booktitle={Proceedings of the AAAI conference on artificial intelligence},
  volume={37},
  number={9},
  pages={11121--11128},
  year={2023}
}

@misc{itransformer,
      title={iTransformer: Inverted Transformers Are Effective for Time Series Forecasting}, 
      author={Yong Liu and Tengge Hu and Haoran Zhang and Haixu Wu and Shiyu Wang and Lintao Ma and Mingsheng Long},
      year={2024},
      eprint={2310.06625},
      archivePrefix={arXiv},
      primaryClass={cs.LG},
      url={https://arxiv.org/abs/2310.06625}, 
}

@inproceedings{fedformer,
  title={Fedformer: Frequency enhanced decomposed transformer for long-term series forecasting},
  author={Zhou, Tian and Ma, Ziqing and Wen, Qingsong and Wang, Xue and Sun, Liang and Jin, Rong},
  booktitle={International conference on machine learning},
  pages={27268--27286},
  year={2022},
  organization={PMLR}
}

@article{chen2024personalized,
  title={Personalized adapter for large meteorology model on devices: Towards weather foundation models},
  author={Chen, Shengchao and Long, Guodong and Jiang, Jing and Zhang, Chengqi},
  journal={Advances in Neural Information Processing Systems},
  volume={37},
  pages={84897--84943},
  year={2024}
}

@article{FeDaL,
  title={FeDaL: Federated Dataset Learning for Time Series Foundation Models},
  author={Chen, Shengchao and Long, Guodong and Jiang, Jing},
  journal={arXiv preprint arXiv:2508.04045},
  year={2025}
}

@inproceedings{federated_NonIID,
  title={Federated learning on non-iid graphs via structural knowledge sharing},
  author={Tan, Yue and Liu, Yixin and Long, Guodong and Jiang, Jing and Lu, Qinghua and Zhang, Chengqi},
  booktitle={Proceedings of the AAAI conference on artificial intelligence},
  volume={37},
  number={8},
  pages={9953--9961},
  year={2023}
}

@article{tan2024language,
  title={Are language models actually useful for time series forecasting?},
  author={Tan, Mingtian and Merrill, Mike and Gupta, Vinayak and Althoff, Tim and Hartvigsen, Tom},
  journal={Advances in Neural Information Processing Systems},
  volume={37},
  pages={60162--60191},
  year={2024}
}

@article{GPT4TS,
  title={One fits all: Power general time series analysis by pretrained lm},
  author={Zhou, Tian and Niu, Peisong and Sun, Liang and Jin, Rong and others},
  journal={Advances in neural information processing systems},
  volume={36},
  pages={43322--43355},
  year={2023}
}

@article{tempo,
  title={Tempo: Prompt-based generative pre-trained transformer for time series forecasting},
  author={Cao, Defu and Jia, Furong and Arik, Sercan O and Pfister, Tomas and Zheng, Yixiang and Ye, Wen and Liu, Yan},
  journal={arXiv preprint arXiv:2310.04948},
  year={2023}
}

@article{forecastpfn,
  title={Forecastpfn: Synthetically-trained zero-shot forecasting},
  author={Dooley, Samuel and Khurana, Gurnoor Singh and Mohapatra, Chirag and Naidu, Siddartha V and White, Colin},
  journal={Advances in Neural Information Processing Systems},
  volume={36},
  pages={2403--2426},
  year={2023}
}

@article{woo2024unified,
  title={Unified training of universal time series forecasting transformers},
  author={Woo, Gerald and Liu, Chenghao and Kumar, Akshat and Xiong, Caiming and Savarese, Silvio and Sahoo, Doyen},
  year={2024},
  publisher={PMLR}
}

@article{TimeGPT-1,
  title={TimeGPT-1},
  author={Garza, Azul and Challu, Cristian and Mergenthaler-Canseco, Max},
  journal={arXiv preprint arXiv:2310.03589},
  year={2023}
}

@article{Moment,
  title={Moment: A family of open time-series foundation models},
  author={Goswami, Mononito and Szafer, Konrad and Choudhry, Arjun and Cai, Yifu and Li, Shuo and Dubrawski, Artur},
  journal={arXiv preprint arXiv:2402.03885},
  year={2024}
}

@article{Timer,
  title={Timer: Generative pre-trained transformers are large time series models},
  author={Liu, Yong and Zhang, Haoran and Li, Chenyu and Huang, Xiangdong and Wang, Jianmin and Long, Mingsheng},
  journal={arXiv preprint arXiv:2402.02368},
  year={2024}
}

@article{pytorch,
  title={Pytorch: An imperative style, high-performance deep learning library},
  author={Paszke, Adam and Gross, Sam and Massa, Francisco and Lerer, Adam and Bradbury, James and Chanan, Gregory and Killeen, Trevor and Lin, Zeming and Gimelshein, Natalia and Antiga, Luca and others},
  journal={Advances in neural information processing systems},
  volume={32},
  year={2019}
}

@article{communication,
  title={Communication-efficient federated learning},
  author={Chen, Mingzhe and Shlezinger, Nir and Poor, H Vincent and Eldar, Yonina C and Cui, Shuguang},
  journal={Proceedings of the National Academy of Sciences},
  volume={118},
  number={17},
  pages={e2024789118},
  year={2021},
  publisher={National Academy of Sciences}
}

@article{boue2025deep,
  title={Deep learning for pedestrians: backpropagation in Transformers},
  author={Bou{\'e}, Laurent},
  journal={arXiv preprint arXiv:2512.23329},
  year={2025}
}

@article{deepseek-r1,
  title={DeepSeek-R1 incentivizes reasoning in LLMs through reinforcement learning},
  author={Guo, Daya and Yang, Dejian and Zhang, Haowei and Song, Junxiao and Wang, Peiyi and Zhu, Qihao and Xu, Runxin and Zhang, Ruoyu and Ma, Shirong and Bi, Xiao and others},
  journal={Nature},
  volume={645},
  number={8081},
  pages={633--638},
  year={2025},
  publisher={Nature Publishing Group UK London}
}

@article{kimi-vl,
  title={Kimi-vl technical report},
  author={Team, Kimi and Du, Angang and Yin, Bohong and Xing, Bowei and Qu, Bowen and Wang, Bowen and Chen, Cheng and Zhang, Chenlin and Du, Chenzhuang and Wei, Chu and others},
  journal={arXiv preprint arXiv:2504.07491},
  year={2025}
}

@inproceedings{airformer,
  title={Airformer: Predicting nationwide air quality in china with transformers},
  author={Liang, Yuxuan and Xia, Yutong and Ke, Songyu and Wang, Yiwei and Wen, Qingsong and Zhang, Junbo and Zheng, Yu and Zimmermann, Roger},
  booktitle={Proceedings of the AAAI conference on artificial intelligence},
  volume={37},
  number={12},
  pages={14329--14337},
  year={2023}
}

@article{liu2024moirai,
  title={Moirai-moe: Empowering time series foundation models with sparse mixture of experts},
  author={Liu, Xu and Liu, Juncheng and Woo, Gerald and Aksu, Taha and Liang, Yuxuan and Zimmermann, Roger and Liu, Chenghao and Savarese, Silvio and Xiong, Caiming and Sahoo, Doyen},
  journal={arXiv preprint arXiv:2410.10469},
  year={2024}
}

@misc{foundation_time_series_survey,
      title={Foundation Models for Time Series: A Survey}, 
      author={Siva Rama Krishna Kottapalli and Karthik Hubli and Sandeep Chandrashekhara and Garima Jain and Sunayana Hubli and Gayathri Botla and Ramesh Doddaiah},
      year={2025},
      eprint={2504.04011},
      archivePrefix={arXiv},
      primaryClass={cs.LG},
      url={https://arxiv.org/abs/2504.04011}, 
}

@misc{koh2020conceptbottleneckmodels,
      title={Concept Bottleneck Models}, 
      author={Pang Wei Koh and Thao Nguyen and Yew Siang Tang and Stephen Mussmann and Emma Pierson and Been Kim and Percy Liang},
      year={2020},
      eprint={2007.04612},
      archivePrefix={arXiv},
      primaryClass={cs.LG},
      url={https://arxiv.org/abs/2007.04612}, 
}

@article{scale_laws_LLM,
  title={Scaling laws for neural language models},
  author={Kaplan, Jared and McCandlish, Sam and Henighan, Tom and Brown, Tom B and Chess, Benjamin and Child, Rewon and Gray, Scott and Radford, Alec and Wu, Jeffrey and Amodei, Dario},
  journal={arXiv preprint arXiv:2001.08361},
  year={2020}
}

@article{scale_laws,
  title={Towards neural scaling laws for time series foundation models},
  author={Yao, Qingren and Yang, Chao-Han Huck and Jiang, Renhe and Liang, Yuxuan and Jin, Ming and Pan, Shirui},
  journal={arXiv preprint arXiv:2410.12360},
  year={2024}
}

@article{scale_laws2,
  title={Scaling law for time series forecasting},
  author={Shi, Jingzhe and Ma, Qinwei and Ma, Huan and Li, Lei},
  journal={Advances in Neural Information Processing Systems},
  volume={37},
  pages={83314--83344},
  year={2024}
}

@inproceedings{liu2025personalized,
  title={Personalized federated learning for spatio-temporal forecasting: A dual semantic alignment-based contrastive approach},
  author={Liu, Qingxiang and Sun, Sheng and Liang, Yuxuan and Liu, Min and Xue, Jingjing},
  booktitle={Proceedings of the AAAI Conference on Artificial Intelligence},
  volume={39},
  number={11},
  pages={12192--12200},
  year={2025}
}

@article{FedRAP,
  title={Federated recommendation with additive personalization},
  author={Li, Zhiwei and Long, Guodong and Zhou, Tianyi},
  journal={arXiv preprint arXiv:2301.09109},
  year={2023}
}

@article{FedPA,
  title={Federated adaptation for foundation model-based recommendations},
  author={Zhang, Chunxu and Long, Guodong and Guo, Hongkuan and Fang, Xiao and Song, Yang and Liu, Zhaojie and Zhou, Guorui and Zhang, Zijian and Liu, Yang and Yang, Bo},
  journal={arXiv preprint arXiv:2405.04840},
  year={2024}
}

@article{t-SNE,
  title={Visualizing data using t-SNE},
  author={Maaten, Laurens van der and Hinton, Geoffrey},
  journal={Journal of machine learning research},
  volume={9},
  number={Nov},
  pages={2579--2605},
  year={2008}
}

@article{chen2023federated,
  title={Federated prompt learning for weather foundation models on devices},
  author={Chen, Shengchao and Long, Guodong and Shen, Tao and Jiang, Jing and Zhang, Chengqi},
  journal={arXiv preprint arXiv:2305.14244},
  year={2023}
}

@incollection{huber1992robust,
  title={Robust estimation of a location parameter},
  author={Huber, Peter J},
  booktitle={Breakthroughs in statistics: Methodology and distribution},
  pages={492--518},
  year={1992},
  publisher={Springer}
}

@article{STD2Vformer,
  title={STD2Vformer: A Free-Form Spatiotemporal Forecasting Model},
  author={Deng, Liwei and Wang, Hao and Tan, Junhao and Niu, Xinhe and He, Yuxin and Zhang, Shiyao and He, Zhihai},
  journal={IEEE Transactions on Industrial Informatics},
  year={2026},
  publisher={IEEE}
}
\bibliographystyle{icml2026}

%%%%%%%%%%%%%%%%%%%%%%%%%%%%%%%%%%%%%%%%%%%%%%%%%%%%%%%%%%%%%%%%%%%%%%%%%%%%%%%
%%%%%%%%%%%%%%%%%%%%%%%%%%%%%%%%%%%%%%%%%%%%%%%%%%%%%%%%%%%%%%%%%%%%%%%%%%%%%%%
% APPENDIX
%%%%%%%%%%%%%%%%%%%%%%%%%%%%%%%%%%%%%%%%%%%%%%%%%%%%%%%%%%%%%%%%%%%%%%%%%%%%%%%
%%%%%%%%%%%%%%%%%%%%%%%%%%%%%%%%%%%%%%%%%%%%%%%%%%%%%%%%%%%%%%%%%%%%%%%%%%%%%%%
\newpage
\appendix
% \onecolumn

\begin{algorithm}[thbp]
   \caption{Implementation of \textsc{FeDPM}}
   \label{alg:fl_vqvae}
\begin{algorithmic}[1]
   \STATE {\bfseries ServerExecute:}
   \STATE Initialize global memories $\{\pmb{P}_{G,1},\dots,\pmb{P}_{G,N}\}$ for each domain randomly
   \FOR{round $t = 1, 2, \dots, T$}
      \FOR{domain $n \in \{1, \dots, N\}$ {\bfseries in parallel}}
         \STATE $\pmb{P}_n, \{\mathrm{Freq}(\pmb{e})\}_{\pmb{e}\in \pmb{P}_n} \leftarrow \text{ClientUpdate}(n, \pmb{P}_{G,n})$
      \ENDFOR
      
      \STATE \textcolor{gray}{\textit{// Cross-Domain Memory Alignment }}
      \STATE Compute cross-domain similarity matrix $\pmb{\mathcal{S}}$ via Eq.~\eqref{eq:cos}
      \STATE Construct graph edges where similarity $> \delta$ and perform BFS to obtain cluster set $\mathcal{K}$
      \STATE Compute aggregated centroid $\pmb{e}_s$ for each cluster via Eq.~\eqref{eq:cluster}
      
      \STATE \textcolor{gray}{\textit{ // Global Consensus Selection}}
      \STATE Set max global capacity $M_g \leftarrow \lfloor \gamma M \rfloor$
      \STATE Determine shared count $K \leftarrow \min(|\mathcal{K}|, M_g)$
      \STATE $\pmb{P}_{S} \leftarrow$ Select top-$K$ centroids $\{\pmb{e}_s\}$ with largest cluster cardinality
      
      \STATE  \textcolor{gray}{\textit{// Personalized Prototypes Completion}}
      \FOR{domain $n \in \{1, \dots, N\}$}
         \STATE Identify unclustered set $\mathcal{U}_n$ for domain $n$
         \STATE Calculate utility-diversity score $\mathcal{V}(\pmb{e})$ for each $\pmb{e} \in \mathcal{U}_n$ via Eq.~\eqref{eq:score}
         \STATE $\pmb{P}_{p,n} \leftarrow$ Select top-$(M-K)$ vectors from $\mathcal{U}_n$ with highest utility-diversity scores
         \STATE $\pmb{P}_{G,n} \leftarrow \pmb{P}_{S} \cup \pmb{P}_{p,n}$
      \ENDFOR
   \ENDFOR

   \STATE {\bfseries ClientUpdate($n, \pmb{P}_{G,n}$):}
   \STATE Initialize local memory $\pmb{P}_n$ with $\pmb{P}_{G,n}$
   \STATE Initialize frequencies $\mathrm{Freq}(\pmb{e}) \leftarrow 0$ for all $\pmb{e} \in \pmb{P}_n$
   \FOR{epoch $e$ from $1$ to $E$}
      \FOR{$(\pmb{X}_n,\pmb{Y}_n ) \in \mathcal{D}_n$}
        \FOR{channel $\ell \in \{1, \dots, c_n\}$ {\bfseries in parallel}}
             \STATE $\hat{\pmb{X}}_{n,S} \leftarrow \text{Patch}(\text{Normalize}(\pmb{x}_n))$ 
             \STATE $\pmb{Z}_n \leftarrow \mathcal{M}_{n,\mathcal{E}}(\hat{\pmb{X}}_{n,S})$ 
             \STATE $\hat{\pmb{Z}}_{n} \leftarrow \text{PMR}(\pmb{Z}_n, \pmb{P}_n)$ with Eq.~\eqref{eq:vq_strat}
             \STATE $\hat{\pmb{H}}_n \leftarrow \mathcal{M}_{n,\mathcal{D}}(\hat{\pmb{Z}}_{n})$
             \STATE $\hat{\pmb{y}}_n \leftarrow \text{De-Normalize}(\text{De-Patch}(\hat{\pmb{H}}_n))$ 
            \ENDFOR
        \STATE Update $\pmb{P}_n$ via gradient descent
        \STATE Update usage frequencies $\mathrm{Freq}(\pmb{e})$ for each codevectors
        \STATE Update Encoder and Decoder parameters via gradient descent
      \ENDFOR
   \ENDFOR
   \STATE {\bfseries Return} $\pmb{P}_n, \{\mathrm{Freq}(\pmb{e})\}_{\pmb{e}\in \pmb{P}_n}$ to server
\end{algorithmic}
\end{algorithm}

\begin{table}[ht]
    \centering
    \caption{Summary of Notations used in \textsc{FeDPM}.}
    \label{tab:notations}
    \resizebox{\columnwidth}{!}{% Resize to fit column if needed
    \begin{tabular}{l l}
        \toprule
        \textbf{Notation} & \textbf{Description} \\
        \midrule
        \multicolumn{2}{c}{\textit{Problem Definition \& Data}} \\
        \midrule
        $N$ & Number of domains (clients) \\
        $n$ & Index of the domain, $n \in \{1, \dots, N\}$ \\
        $\mathcal{D}_n$ & Local dataset of domain $n$ \\
        $\pmb{X}_n$ & Input time series sequence, $\pmb{X}_n \in \mathbb{R}^{L_n \times c_n}$ \\
        $\pmb{Y}_n$ & Ground truth (future) sequence, $\pmb{Y}_n \in \mathbb{R}^{F_n \times c_n}$ \\
        $L_n, F_n$ & Look-back window and prediction horizon for domain $n$ \\
        $c_n$ & Number of channels (variables) in domain $n$ \\
        
        \midrule
        \multicolumn{2}{c}{\textit{Model Architecture} (Default: domain $n$, channel-level)} \\
        \midrule
        $\mathcal{M}_{n, \mathcal{E}}$ & Encoder module for domain $n$ \\
        $\mathcal{M}_{n, \mathcal{D}}$ & Decoder module for domain $n$ \\
        $\pmb{Z}_n$ & Latent representation \\
        $\hat{\pmb{Z}}_n$ & Quantized latent representation after PMR \\
        $\hat{\pmb{H}}_n$ & Output of the decoder \\
        $\hat{\pmb{y}}_n$ & Final forecasted time series \\
        $sg(\cdot)$ & Stop-gradient operator \\
        
        \midrule
        \multicolumn{2}{c}{\textit{Prototype \& Memory}} \\
        \midrule
        $\pmb{P}_n$ & Local Memory for domain $n$ \\
        $\pmb{P}_{G,n}$ & Global Memory for domain $n$ \\
        $\pmb{P}_s$ & Shared Prototypes (Global Consensus) \\
        $\pmb{P}_{p,n}$ & Personalized Prototypes for domain $n$ \\
        $M$ & Memory size (number of prototype vectors) \\
        $D$ & Dimension of prototype vectors \\
        $\pmb{e}_{n,m}$ & The $m$-th prototype vector in domain $n$'s memory \\
        $\mathcal{K}$ & Set of clusters formed during aggregation \\
        $\delta$ & Threshold for cross-domain cosine similarity \\
        $\gamma$ & Ratio controlling the maximum global consensus capacity \\
        \bottomrule
    \end{tabular}
    }
\end{table}

\begin{table*}[thbp]
\centering
\caption{Results of ablation experiments on Time-FFM \cite{time-ffm}. \textbf{Bold} denotes the best performance, and \underline{underlined} results indicate improvements over LLM-based baselines.}
\label{tab:Ablation_Time_FFM}
\resizebox{2\columnwidth}{!}{%
\begin{tabular}{cc|cc|cc|cc|cc|cc|cc|cc}
\toprule
\multirow{2}{*}{Method}      & Dataset & \multicolumn{2}{c|}{ETTh1}             & \multicolumn{2}{c|}{ETTh2}       & \multicolumn{2}{c|}{ETTm1}             & \multicolumn{2}{c|}{ETTm2}                   & \multicolumn{2}{c|}{Electricity}             & \multicolumn{2}{c|}{Exchange}                & \multicolumn{2}{c}{Weather}                 \\
                             & Length  & MSE                  & MAE            & MSE            & MAE            & MSE                  & MAE            & MSE                  & MAE                  & MSE                  & MAE                  & MSE                  & MAE                  & MSE                  & MAE                  \\
                             \midrule
\multirow{4}{*}{LLMs}        & 96      & 0.406                & \textbf{0.404} & \textbf{0.293} & \textbf{0.341} & 0.357                & \textbf{0.373} & 0.180                & 0.264                & 0.207                & 0.295                & 0.087                & 0.203                & 0.198                & 0.238                \\
                             & 192     & 0.460                & \textbf{0.434} & \textbf{0.372} & \textbf{0.391} & 0.399                & \textbf{0.393} & 0.245                & 0.304                & 0.209                & 0.300                & 0.187                & 0.304                & 0.242                & 0.273                \\
                             & 336     & 0.504                & \textbf{0.453} & \textbf{0.413} & \textbf{0.426} & 0.428                & \textbf{0.411} & \textbf{0.306}       & \textbf{0.343}       & 0.225                & 0.316                & 0.341                & 0.421                & 0.295                & 0.310                \\
                             & 720     & \textbf{0.495}       & \textbf{0.466} & \textbf{0.419} & \textbf{0.440} & 0.490                & \textbf{0.444} & \textbf{0.404}       & \textbf{0.398}       & 0.264                & 0.344                & \textbf{0.891}       & \textbf{0.714}       & 0.370                & 0.358                \\
                             \midrule
\multirow{4}{*}{Transformer} & 96      & \underline{\textbf{0.391}} & 0.409          & 0.307          & 0.354          & \underline{\textbf{0.338}} & 0.373          & 0.185                & 0.272                & \underline{\textbf{0.181}} & \underline{\textbf{0.270}} & \underline{\textbf{0.082}} & \underline{\textbf{0.202}} & \underline{\textbf{0.179}} & \underline{\textbf{0.224}} \\
                             & 192     & \underline{\textbf{0.445}} & 0.442          & 0.384          & 0.406          & \underline{\textbf{0.387}} & 0.398          & 0.258                & 0.320                & \underline{\textbf{0.185}} & \underline{\textbf{0.275}} & \underline{\textbf{0.173}} & \underline{\textbf{0.298}} & \underline{\textbf{0.225}} & \underline{0.262}          \\
                             & 336     & \underline{0.497}          & 0.471          & 0.427          & 0.440          & \underline{\textbf{0.420}} & 0.420          & 0.327                & 0.365                & \underline{\textbf{0.200}} & \underline{\textbf{0.290}} & \underline{\textbf{0.323}} & \underline{\textbf{0.411}} & \underline{\textbf{0.280}} & \underline{0.301}          \\
                             & 720     & 0.537                & 0.511          & 0.447          & 0.460          & \underline{\textbf{0.484}} & 0.458          & 0.429                & 0.425                & \underline{\textbf{0.240}} & \underline{\textbf{0.322}} & 0.947                & 0.728                & \underline{\textbf{0.355}}          & \underline{0.350}          \\
                             \midrule
\multirow{4}{*}{FC}          & 96      & \underline{0.404}          & 0.413          & 0.305          & 0.351          & 0.376                & 0.387          & \underline{\textbf{0.177}} & \underline{\textbf{0.260}} & 0.225                & 0.319                & 0.088                & 0.206                & \underline{0.182}          & \underline{0.224}          \\
                             & 192     & \underline{0.451}          & 0.441          & 0.385          & 0.400          & 0.410                & 0.405          & \underline{\textbf{0.243}} & \underline{\textbf{0.304}} & 0.226                & 0.321                & 0.189                & 0.305                & \underline{0.226}          & \underline{\textbf{0.261}} \\
                             & 336     & \underline{\textbf{0.488}} & 0.460          & 0.424          & 0.430          & 0.437                & 0.423          & 0.309                & 0.346                & 0.239                & 0.333                & 0.342                & \underline{0.421}          & \underline{0.280}          & \underline{\textbf{0.299}} \\
                             & 720     & 0.502                & 0.487          & 0.429          & 0.444          & 0.502                & 0.459          & 0.414                & 0.407                & 0.279                & 0.361                & 0.917                & 0.717                & \underline{\textbf{0.355}} & \underline{\textbf{0.348}} \\
                             \bottomrule
\end{tabular}%
}
\end{table*}

\section{Ablation Experiment Conducted on Time-FFM} \label{Ablation_Time-FFM}
% \begin{figure*}[tbh]
%     \centering
%     \includegraphics[width=0.7\linewidth]{Figure/Experiment/Concatenated_Results.pdf}
%     \caption{Ablation experiments conducted on Time-FFM \cite{time-ffm}}
%     \label{fig:motivation_Full}
% \end{figure*}

To thoroughly address the question \textit{whether pretrained LLMs can actually generalize to time series data in FL setting?}, we conduct an ablation study on Time-FFM \cite{time-ffm} under the full-shot settings.
% three experimental settings: full-shot, zero-shot, and few-shot (5\% training data).
Following the original design of Time-FFM, we adopt a frozen GPT-2 as the LLM backbone, which is also a commonly used choice in time series forecasting with LLMs \cite{unitime,GPT4TS,time-llm,llm4ts,time-ffm}. We then replace the frozen LLMs backbone with two lightweight, fully trainable alternatives:
(i) two Transformer encoder layers, and
(ii) two fully connected (FC) layers.
Experimental results indicate that replacing the frozen LLM backbone with a fully trainable native time series model yields lower MSE in 20 out of 28 evaluated cases (\textbf{71.43\%}) under the full-shot setting with only \textbf{10.1\%} parameters on average.
% Similar performance gains are observed in 4 out of 8 cases (\textbf{50}\%) for zero-shot forecasting and 11 out of 12 cases (\textbf{91.66}\%) for the few-shot setting.

These results indicate that the cross-modal alignment capability of current LLMs backbones for time series modeling remains limited in federated environments. This observation is consistent with the findings of \cite{tan2024language}, which reach a similar conclusion under centralized training settings.

\section{Training Process} \label{appendix:Training_Process}
The overall training procedure of \textsc{FeDPM} is summarized in Algorithm \ref{alg:fl_vqvae}. The framework operates in a federated manner, alternating between learning of Local Prototypical Memory Priors on domain-specific clients and Cross-Domain Memory Updates on the server. The process consists of four key phases: Local Prototypical Memory Priors, Global Consensus Extraction via Cross-Domain Memory Alignment, and Personalized Prototype Completion.

\textbf{Local Prototypical Memory Priors.} At the beginning of each round $t$, the server distributes the personalized global memory $\pmb{P}_{G,n}$ to each domain $n$. Each client initializes its local memory $\pmb{P}_n$ and resets the prototype usage frequencies $\mathrm{Freq}(\pmb{e})$. During the local training epoch, the client processes multi-channel inputs $\pmb{X}_n$. As detailed in lines 27--38, the input patches are normalized and encoded into latent vector $\pmb{Z}_n$ via the encoder $\mathcal{M}_{n,\mathcal{E}}$. These vectors undergo Prototypical Memory Retrieval (via Eq.~\eqref{eq:vq_strat}) using the local memory, followed by prediction via the decoder $\mathcal{M}_{n,\mathcal{D}}$. Crucially, alongside gradient-based updates for the memory and model parameters, the client tracks the cumulative usage frequency of each prototype. Upon completion, the updated memory $\pmb{P}_n$ and the corresponding frequency statistics $\{\mathrm{Freq}(\pmb{e})\}_{\pmb{e}\in \pmb{P}_n}$ are uploaded to the server.

\textbf{Cross-Domain Memory Alignment.} The server aggregates the uploaded memories to identify shared semantic patterns across domains. Instead of simple averaging, we employ a graph-theoretic approach. First, we compute a cross-domain similarity matrix $\pmb{\mathcal{S}}$ (via Eq.~\eqref{eq:cos}) among all uploaded prototypes. A graph is constructed by establishing edges between vectors where the similarity exceeds a threshold $\delta$. By performing Breadth-First Search (BFS) \cite{BFS} on this graph, we obtain a set of clusters $\mathcal{K}$, where each cluster represents a semantic concept \cite{koh2020conceptbottleneckmodels} shared by multiple domains.

\textbf{Global Consensus Extraction.} To form the global consensus, we compute the aggregated centroid $\pmb{e}_s$ for each cluster (via Eq.~\eqref{eq:cluster}). We then determine a shared capacity $K = \min(|\mathcal{K}|, \lfloor \gamma M \rfloor)$, where $\gamma$ controls the maximum ratio of global consensus. The server selects the top-$K$ centroids associated with the largest cluster cardinalities to form the shared prototype subset $\pmb{P}_{S}$. This ensures that the global prototype captures the most prevalent cross-domain consensus.

\textbf{Personalized Prototypes Completion.} To preserve domain-specific characteristics, the remaining capacity of the memory is filled via a personalized completion strategy. For each domain $n$, the server identifies the unclustered set $\mathcal{U}_n$ containing vectors that were not selected for the global consensus. We calculate a utility-diversity score $\mathcal{V}(\pmb{e})$ for each candidate vector in $\mathcal{U}_n$ (via Eq.~\eqref{eq:score}), which typically balances frequency and representational quality. The top-$(M-K)$ vectors with the highest scores are selected as the personalized subset $\pmb{P}_{p,n}$ for domain $n$. Finally, the new global memory for domain $n$ for the next round is assembled as the union of the shared consensus and the personalized subset: $\pmb{P}_{G,n} \leftarrow \pmb{P}_{S} \cup \pmb{P}_{p,n}$. This mechanism allows \textsc{FeDPM} to dynamically balance common knowledge sharing with domain-specific adaptation.

\section{Experimental Details}
\label{sec:Experimental_Details}
\paragraph{Implementation Details.}
We adopt the Adam optimizer with a learning rate of $1\times10^{-5}$ for all experiments. The look-back window length is fixed to $L_n=96$ for all datasets, while the prediction horizon $F_i$ is set to $\{96, 192, 336, 720\}$. The number of local training epochs is set to $E=5$ for all domains, and the total number of federated communication rounds is $T=100$. We apply early stopping with a patience of 10 rounds based on the validation loss.
At each communication round, we compute the average validation loss across all clients. The model checkpoint corresponding to the round with the lowest validation loss is selected and evaluated on the test set. All models are implemented in PyTorch \cite{pytorch}. All experiments are conducted on NVIDIA RTX 5090 GPUs, except for the model efficiency experiment, which are performed on NVIDIA A100-80G GPUs.

\paragraph{Hyperparameter Settings.}
Both the encoder and decoder adopt the standard Transformer architecture \cite{transformer}. Unless otherwise specified, the memory size is set to $M=256$, and the dimensionality of each prototype is $D=64$. The maximum proportion of shared clusters is controlled by $\gamma$, which is set to $0.95$ by default. Following the standard setting in \cite{vqvae}, we set the relative learning rate between the encoder and the memory to $\beta=0.25$ for all experiments.
In addition, both the stride length and patch length are fixed to $S_n=4$ across all domains, and the similarity threshold $\delta$ is set to $0.7$. We conduct a comprehensive hyperparameter sensitivity analysis in Appendix~\ref{sec:hyperparameter_sensitivity}. Further implementation details and hyperparameter configurations are provided in the released code.

\paragraph{Baseline Implementation.}
All baseline models are reproduced using the official implementations released by the authors, with their recommended hyperparameter settings.
For FL-iTransformer and FL-PatchTST, we adapt the corresponding expert models to the federated learning setting by sharing the model parameters across clients via FedAvg~\cite{fedavg}.
For Cen-PatchTST, following UniTime~\cite{unitime}, we convert PatchTST into a centralized time-series foundation model by pretraining it on aggregated datasets from all domains.
For FFTS~\cite{FFTS}, the original paper pretrains the model using additional external datasets. To ensure a fair comparison, we re-implement FFTS under a controlled setting, where the pretraining stage is restricted to the same seven datasets used in our experiments—ETTh1, ETTh2, ETTm1, ETTm2, Electricity, Weather, and Exchange—and the model is further fine-tuned for only 5 epochs.

\begin{table}[th]
  \scriptsize
  \centering
  \setlength{\tabcolsep}{2pt}  % 减少列间距(默认6pt)
\renewcommand{\arraystretch}{1}  % 压缩行高
  \caption{Detailed descriptions of datasets. The dataset size is organized in (training, validation, test).}
  \resizebox{1\columnwidth}{!}{%
    \begin{tabular}{ccccccc}
    \toprule
    Dataset & $c_n$    & Dataset Size & Batch Size  & Frequency & Application Domain \\
    \midrule
    ETTh1 & 7     & (8545, 2881, 2881) & 32      & 1 hour & Electrical Asset Monitoring \\
    ETTh2 & 7     & (8545, 2881, 2881) & 32     & 1 hour & Electrical Asset Monitoring \\
    ETTm1 & 7     & (34465, 11521, 11521) & 64      & 15 minutes & Electrical Asset Monitoring \\
    ETTm2 & 7     & (34465, 11521, 11521) & 64      & 15 minutes & Electrical Asset Monitoring \\
    Electricity & 321   & (18317, 2633, 5261) & 24       & 1 hour & Energy Consumption \\
    Weather & 21    & (36792, 5271, 10540) & 64     & 10 minutes & Weather Forecasting \\
    Exchange &  8    & (5120, 665, 1422) & 24     & 1 day & International Trade \\
    \bottomrule
    \end{tabular}%
    }
  \label{dataset}%
\end{table}%

\begin{figure*}[!t] 
    \centering
    % --- 第一行：3张图 (使用 \hfill 撑满整行) ---
    \begin{subfigure}[b]{0.32\textwidth}
        \centering
        \includegraphics[width=\linewidth]{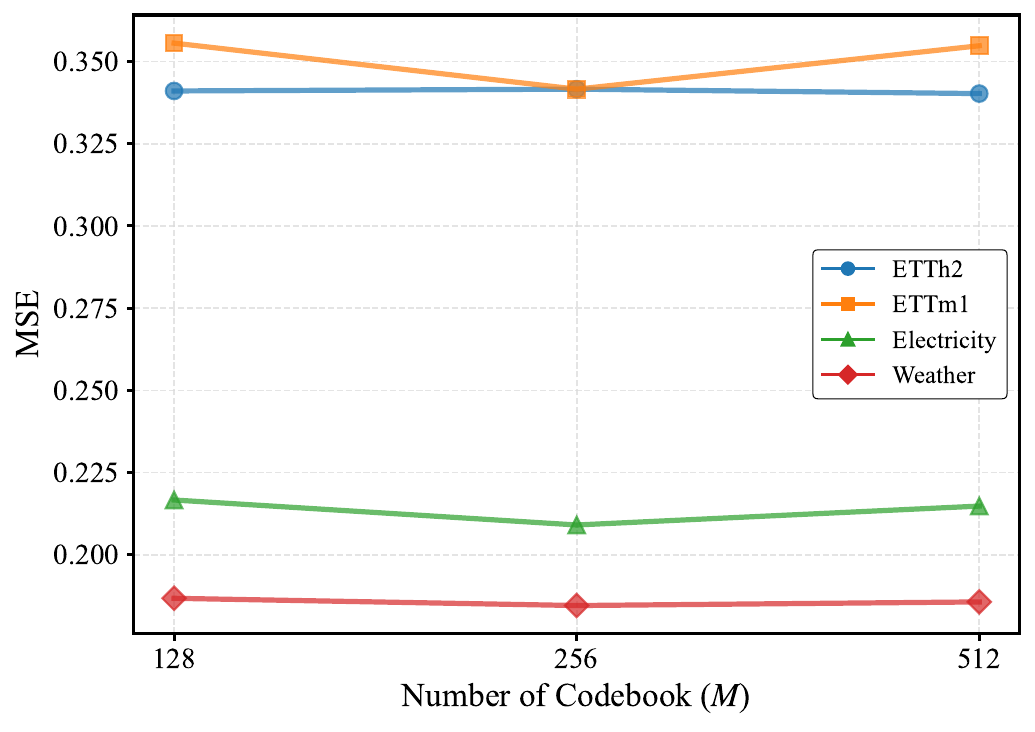}
        \caption{Codebook Size $M$}
        \label{fig:sens_codebook}
    \end{subfigure}
    \hfill % 这里的 hfill 会把三张图撑开到页面两边
    \begin{subfigure}[b]{0.32\textwidth}
        \centering
        \includegraphics[width=\linewidth]{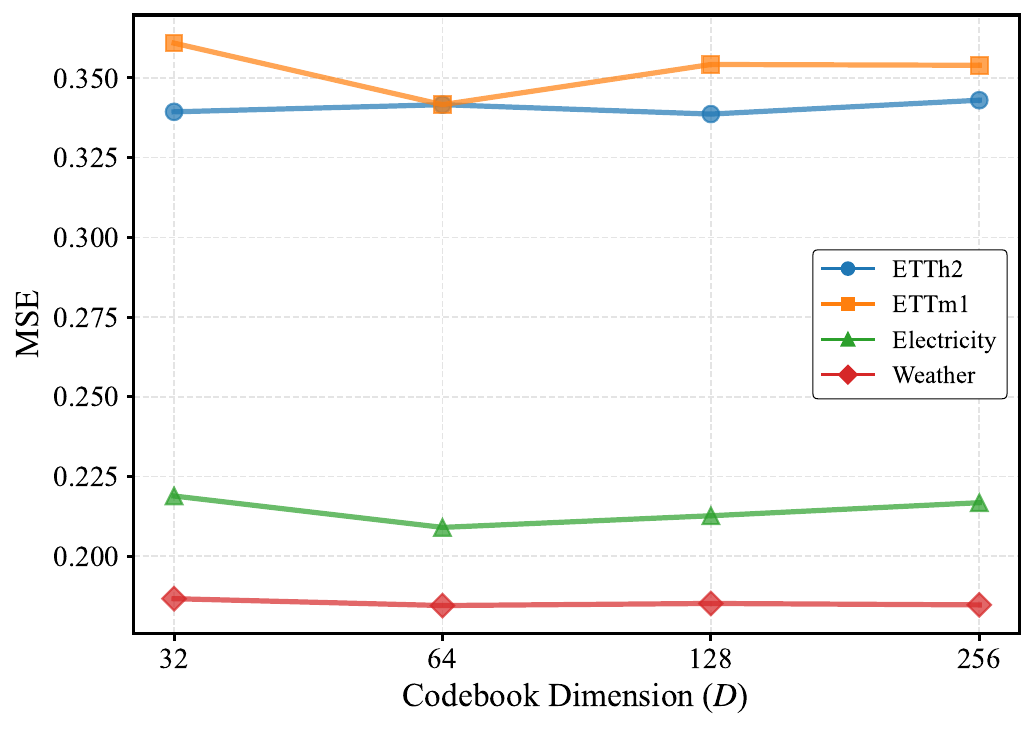}
        \caption{Dimension $D$}
        \label{fig:sens_patch}
    \end{subfigure}
    \hfill
    \begin{subfigure}[b]{0.32\textwidth}
        \centering
        \includegraphics[width=\linewidth]{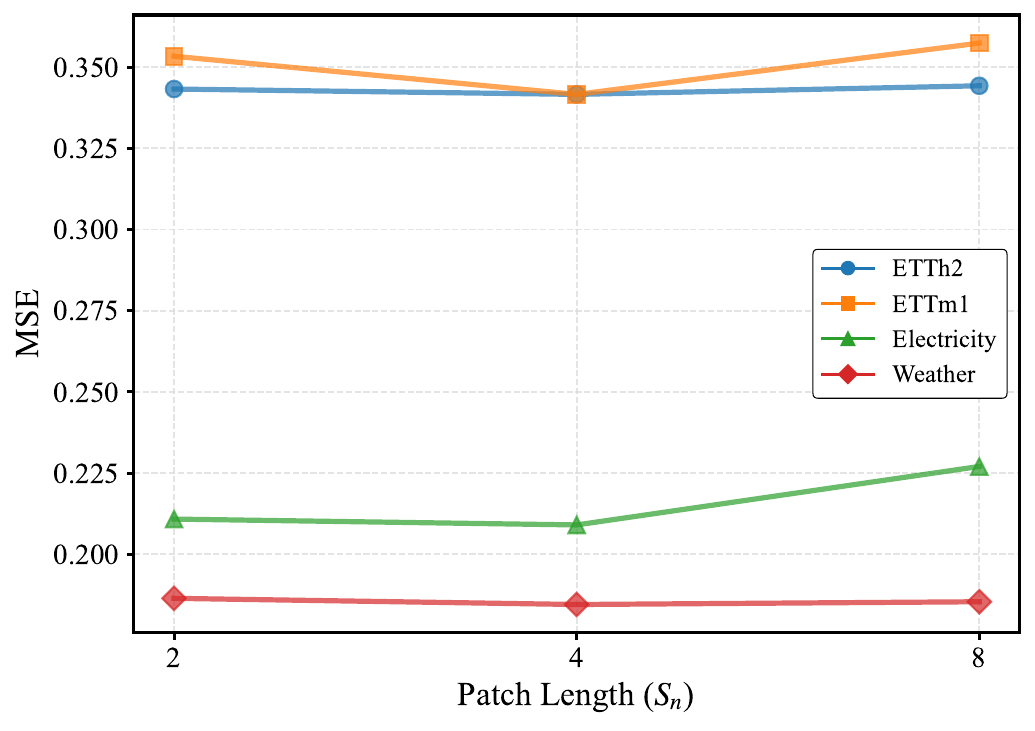}
        \caption{Patch Length $S_n$}
        \label{fig:sens_thresh}
    \end{subfigure}
    
    % --- 强制换行并增加垂直间距 ---
    \par\vspace{1em} 
    
    % --- 第二行：2张图 (关键修改) ---
    % 注意：这里不要用 \hfill，否则会一左一右。
    % 使用 \centering (已在顶部定义) + \hspace 即可让它们在这一行居中。
    \begin{subfigure}[b]{0.4\textwidth}
        \centering
        \includegraphics[width=\linewidth]{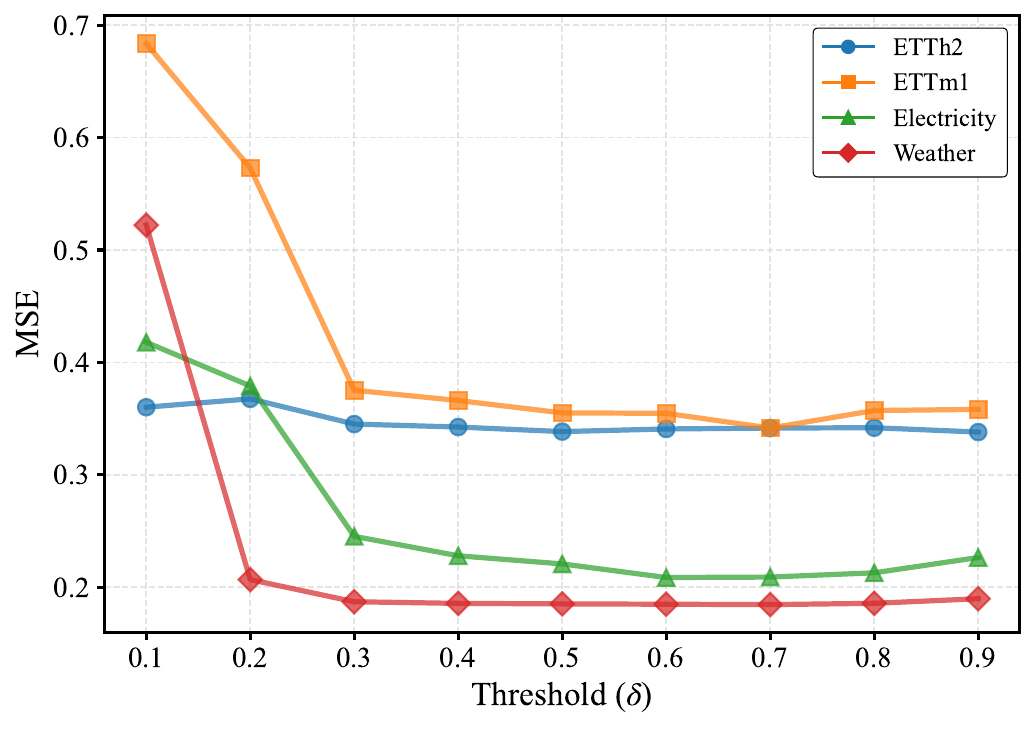}
        \caption{Threshold $\delta$}
        \label{fig:sens_dim}
    \end{subfigure}
    \hspace{4em} % 【关键】这里用固定间距(如4em或0.05\textwidth)，不要用 \hfill
    \begin{subfigure}[b]{0.4\textwidth}
        \centering
        % 【修正】文件名改为 gamma
        \includegraphics[width=\linewidth]
        {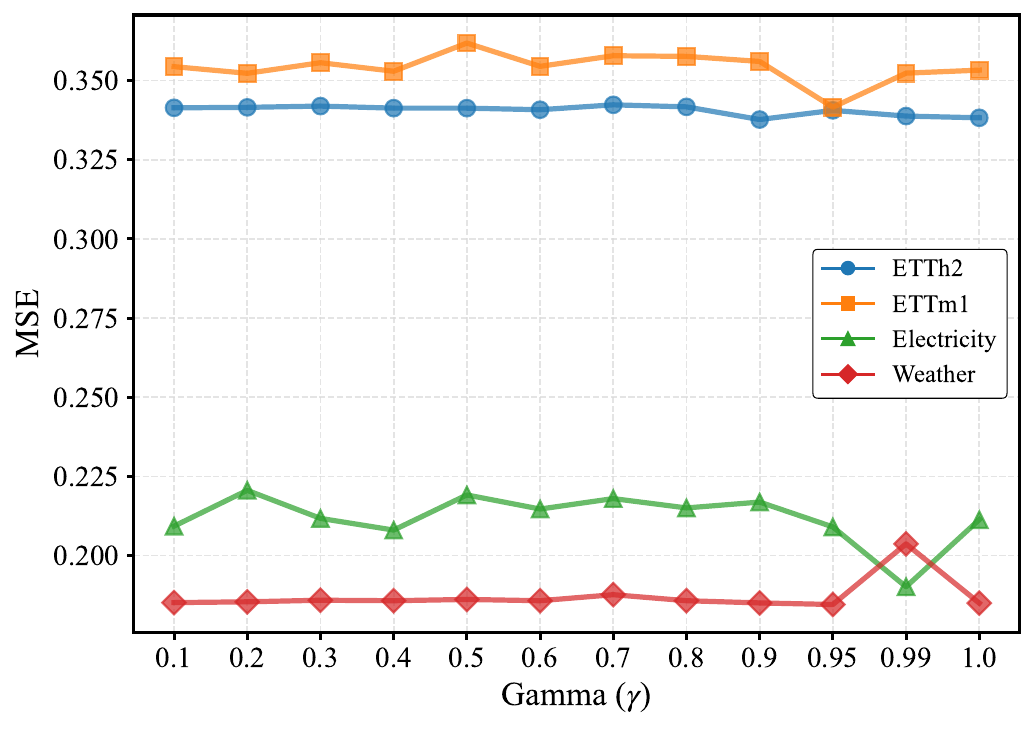} 
        \caption{Gamma $\gamma$}
        \label{fig:sens_gamma}
    \end{subfigure}
    \caption{Hyperparameter Sensitivity Analysis. We evaluate the effects of five key hyperparameters across four datasets under two forecasting horizons, $F_i \in \{96, 192\}$.}
    \label{fig:sensitivity_all}
\end{figure*}

\paragraph{Training Configurations.}
The experimental evaluations are conducted on 7 real-world benchmark datasets which include 4 domains. We present the detailed description of these datasets in Table \ref{dataset}.
For fair comparison, we perform batch division as per \cite{totem}.

\section{Hyperparameter Sensitivity} \label{sec:hyperparameter_sensitivity}
Figure \ref{fig:sensitivity_all} presents the sensitivity analysis for five core hyperparameters: patch length $S_n$, codebook size $M$, dimension $D$, aggregation threshold $\delta$, and the shared ratio $\gamma$. We evaluate these parameters across four benchmarks with prediction lengths of $\{96, 192\}$. Results indicate that the model achieves optimal stability and accuracy with the default settings of $M=256$, $S_n=4$, $D=64$, $\delta=0.7$, and $\gamma=0.95$.

\section{Full Results for Few-Shot Forecasting}
\begin{table*}[thbp]
\centering
\setlength{\tabcolsep}{4pt}  % 减少列间距(默认6pt)
\renewcommand{\arraystretch}{1}  % 压缩行高
\caption{Few-shot results of forecasting performance comparisons. \textbf{Bold:} the best over all types. ‘-’ means time series data is not sufficient to constitute a training set.} 
\label{tab:few_shot_full_result}
\resizebox{2\columnwidth}{!}{%
\begin{tabular}{cc|cc|cc|cc|cc|cc|cc|cc|cc}
\toprule
\rowcolor{blue!15}
    \multicolumn{18}{c}{{\bf Few-shot Long-term Forecasting (5\%)}}\\  
\midrule
\multicolumn{2}{c|}{Type}           & \multicolumn{10}{c|}{FL-FM}                                                                                                                                               & \multicolumn{6}{c}{Cen-FM}                                                                     \\
\midrule
\multicolumn{2}{c|}{Method}         & \multicolumn{2}{c|}{\textsc{FeDPM}}        & \multicolumn{2}{c|}{Time-FFM} & \multicolumn{2}{c|}{FFTS}        & \multicolumn{2}{c|}{FL-iTransformer} & \multicolumn{2}{c|}{FL-PatchTST} & \multicolumn{2}{c|}{TOTEM} & \multicolumn{2}{c|}{UniTime}     & \multicolumn{2}{c}{Cen-PatchTST} \\
\midrule
\multicolumn{2}{c}{Metric}         & MSE            & MAE            & MSE           & MAE          & MSE            & MAE            & MSE              & MAE              & MSE            & MAE            & MSE         & MAE         & MSE            & MAE            & MSE             & MAE            \\
\midrule
\multirow{4}{*}{\rotatebox{90}{ETTm1}}       & 96  & \textbf{0.472} & \textbf{0.441} & 0.515         & 0.459        & 0.538          & 0.492          & 0.879            & 0.601            & 0.866          & 0.548          & 0.928       & 0.693       & 0.576          & 0.498          & 0.559           & 0.477          \\
                             & 192 & \textbf{0.499} & \textbf{0.461} & 0.550         & 0.478        & 0.565          & 0.507          & 1.093            & 0.671            & 0.869          & 0.558          & 0.905       & 0.691       & 0.617          & 0.520          & 0.588           & 0.493          \\
                             & 336 & \textbf{0.558} & \textbf{0.490} & 0.563         & 0.491        & 0.619          & 0.531          & 1.112            & 0.690            & 0.839          & 0.562          & 0.894       & 0.697       & 0.633          & 0.533          & 0.587           & 0.497          \\
                             & 720 & \textbf{0.624} & \textbf{0.529} & 0.641         & 0.536        & 0.729          & 0.601          & 1.235            & 0.736            & 1.024          & 0.649          & 0.892       & 0.695       & 1.028          & 0.680          & 0.631           & 0.522          \\
                             \midrule
\multirow{4}{*}{\rotatebox{90}{ETTm2}}       & 96  & 0.210          & 0.277          & 0.192         & 0.272        & \textbf{0.128} & \textbf{0.242} & 0.244            & 0.322            & 0.201          & 0.283          & 0.382       & 0.465       & 0.198          & 0.279          & 0.200           & 0.282          \\
                             & 192 & 0.271          & 0.315          & 0.254         & 0.311        & \textbf{0.155} & \textbf{0.266} & 0.336            & 0.374            & 0.261          & 0.314          & 0.559       & 0.557       & 0.266          & 0.323          & 0.260           & 0.318          \\
                             & 336 & 0.328          & 0.351          & 0.312         & 0.346        & \textbf{0.193} & \textbf{0.296} & 0.457            & 0.440            & 0.341          & 0.365          & 0.719       & 0.629       & 0.337          & 0.366          & 0.318           & 0.352          \\
                             & 720 & 0.431          & 0.409          & 0.415         & 0.403        & \textbf{0.254} & \textbf{0.339} & 0.822            & 0.584            & 0.512          & 0.454          & 0.872       & 0.688       & 0.453          & 0.430          & 0.419           & 0.407          \\
                             \midrule
\multirow{4}{*}{\rotatebox{90}{Electricity}} & 96  & 0.230          & 0.321          & 0.312         & 0.394        & 0.374          & 0.449          & \textbf{0.195}   & \textbf{0.277}   & 0.241          & 0.342          & 1.025       & 0.822       & 0.281          & 0.371          & 0.295           & 0.379          \\
                             & 192 & 0.232          & 0.325          & 0.305         & 0.391        & 0.360          & 0.440          & \textbf{0.201}   & \textbf{0.285}   & 0.235          & 0.334          & 1.014       & 0.820       & 0.283          & 0.377          & 0.293           & 0.382          \\
                             & 336 & 0.249          & 0.341          & 0.321         & 0.401        & 0.392          & 0.466          & \textbf{0.221}   & \textbf{0.306}   & 0.241          & 0.335          & 1.038       & 0.828       & 0.294          & 0.385          & 0.308           & 0.392          \\
                             & 720 & \textbf{0.279} & \textbf{0.362} & 0.358         & 0.427        & 0.827          & 0.744          & 0.323            & 0.392            & 0.316          & 0.390          & 1.044       & 0.831       & 0.335          & 0.413          & 0.341           & 0.413          \\
                             \midrule
\multirow{4}{*}{\rotatebox{90}{Weather}}     & 96  & \textbf{0.179} & \textbf{0.228} & 0.214         & 0.265        & 0.193          & 0.241          & 0.225            & 0.254            & 0.196          & 0.233          & 0.253       & 0.291       & 0.209          & 0.260          & 0.221           & 0.271          \\
                             & 192 & \textbf{0.223} & \textbf{0.270} & 0.264         & 0.302        & 0.241          & 0.278          & 0.296            & 0.311            & 0.250          & 0.275          & 0.281       & 0.309       & 0.258          & 0.297          & 0.271           & 0.308          \\
                             & 336 & \textbf{0.279} & \textbf{0.309} & 0.310         & 0.329        & 0.294          & 0.315          & 0.388            & 0.365            & 0.340          & 0.347          & 0.323       & 0.340       & 0.306          & 0.325          & 0.318           & 0.336          \\
                             & 720 & \textbf{0.347} & \textbf{0.352} & 0.381         & 0.374        & 0.372          & 0.367          & 0.510            & 0.431            & 0.416          & 0.388          & 0.361       & 0.366       & 0.380          & 0.371          & 0.391           & 0.382          \\
                             \midrule
\multirow{4}{*}{\rotatebox{90}{Exchange}}    & 96  & \textbf{0.100} & \textbf{0.237} & 0.118         & 0.244        & 0.140          & 0.270          & 0.126            & 0.256            & 0.121          & 0.251          & 1.550       & 1.003       & 0.385          & 0.458          & 0.123           & 0.250          \\
                             & 192 & 0.210          & 0.350          & 0.215         & 0.334        & 0.235          & 0.352          & \textbf{0.205}   & \textbf{0.324}   & 0.240          & 0.357          & 1.688       & 1.049       & 0.498          & 0.528          & 0.220           & 0.337          \\
                             & 336 & -              & -              & -             & -            & -              & -              & -                & -                & -              & -              & -           & -           & -              & -              & -               & -              \\
                             & 720 & -              & -              & -             & -            & -              & -              & -                & -                & -              & -              & -           & -           & -              & -              & -               & -              \\
                             \midrule
\multicolumn{2}{c|}{$1^{st}$ Count}      & \multicolumn{2}{c|}{\textbf{20}} & \multicolumn{2}{c|}{0}        & \multicolumn{2}{c|}{8}           & \multicolumn{2}{c|}{8}               & \multicolumn{2}{c|}{0}           & \multicolumn{2}{c|}{0}      & \multicolumn{2}{c|}{0}           & \multicolumn{2}{c}{0}            \\
\midrule
\rowcolor{blue!15}
    \multicolumn{18}{c}{{\bf Few-shot Long-term Forecasting (10\%)}}\\                              
% \multicolumn{2}{c}{Type}           & \multicolumn{10}{c}{FL-FM}                                                                                                                                               & \multicolumn{6}{c}{Cen-FM}                                                                     \\
% \multicolumn{2}{c}{Method}         & \multicolumn{2}{c}{\textsc{FeDPM}}        & \multicolumn{2}{c}{Time-FFM} & \multicolumn{2}{c}{FFTS}        & \multicolumn{2}{c}{FL-iTransformer} & \multicolumn{2}{c}{FL-PatchTST} & \multicolumn{2}{c}{TOTEM} & \multicolumn{2}{c}{UniTime}     & \multicolumn{2}{c}{Cen-PatchTST} \\
% \multicolumn{2}{c}{Metric}         & MSE            & MAE            & MSE           & MAE          & MSE            & MAE            & MSE              & MAE              & MSE            & MAE            & MSE         & MAE         & MSE            & MAE            & MSE             & MAE            \\
\midrule
\multirow{4}{*}{\rotatebox{90}{ETTm1}}       & 96  & \textbf{0.547} & \textbf{0.468} & 0.571         & 0.481        & 0.575          & 0.512          & 1.050            & 0.640            & 1.041          & 0.583          & 0.829       & 0.613       & 0.582          & 0.485          & 1.136           & 0.672          \\
                             & 192 & \textbf{0.508} & \textbf{0.462} & 0.578         & 0.490        & 0.601          & 0.521          & 1.177            & 0.682            & 0.895          & 0.568          & 0.822       & 0.611       & 0.564          & 0.479          & 1.118           & 0.672          \\
                             & 336 & 0.625          & 0.516          & 0.592         & 0.504        & 0.642          & 0.540          & 1.076            & 0.670            & 1.001          & 0.614          & 0.788       & 0.599       & \textbf{0.578} & \textbf{0.489} & 0.987           & 0.637          \\
                             & 720 & \textbf{0.622} & 0.525          & 0.629         & 0.526        & 0.725          & 0.588          & 1.418            & 0.764            & 1.942          & 0.822          & 0.803       & 0.608       & 0.631          & \textbf{0.523} & 1.044           & 0.666          \\
                             \midrule
\multirow{4}{*}{\rotatebox{90}{ETTm2}}       & 96  & 0.211          & 0.274          & 0.195         & 0.277        & \textbf{0.129} & \textbf{0.245} & 0.218            & 0.294            & 0.194          & 0.272          & 0.260       & 0.350       & 0.192          & 0.274          & 0.255           & 0.329          \\
                             & 192 & 0.267          & 0.311          & 0.256         & 0.313        & \textbf{0.154} & \textbf{0.267} & 0.293            & 0.340            & 0.257          & 0.313          & 0.347       & 0.417       & 0.256          & 0.313          & 0.312           & 0.360          \\
                             & 336 & 0.325          & 0.347          & 0.314         & 0.348        & \textbf{0.188} & \textbf{0.293} & 0.393            & 0.396            & 0.327          & 0.356          & 0.399       & 0.447       & 0.320          & 0.352          & 0.359           & 0.384          \\
                             & 720 & 0.424          & 0.403          & 0.412         & 0.403        & \textbf{0.246} & \textbf{0.333} & 0.587            & 0.480            & 0.437          & 0.417          & 0.514       & 0.509       & 0.429          & 0.413          & 0.465           & 0.440          \\
                             \midrule
\multirow{4}{*}{\rotatebox{90}{Electricity}} & 96  & 0.225          & 0.316          & 0.249         & 0.329        & 0.374          & 0.448          & \textbf{0.184}   & \textbf{0.271}   & 0.246          & 0.351          & 0.946       & 0.792       & 0.236          & 0.327          & 0.344           & 0.416          \\
                             & 192 & 0.228          & 0.322          & 0.247         & 0.330        & 0.359          & 0.436          & \textbf{0.191}   & \textbf{0.277}   & 0.218          & 0.314          & 0.946       & 0.794       & 0.236          & 0.328          & 0.343           & 0.418          \\
                             & 336 & 0.246          & 0.337          & 0.267         & 0.346        & 0.375          & 0.448          & \textbf{0.215}   & \textbf{0.300}   & 0.262          & 0.364          & 0.948       & 0.795       & 0.250          & 0.341          & 0.361           & 0.429          \\
                             & 720 & 0.279          & 0.361          & 0.300         & 0.368        & 0.417          & 0.475          & \textbf{0.265}   & \textbf{0.340}   & 0.282          & 0.362          & 0.956       & 0.800       & 0.295          & 0.371          & 0.399           & 0.453          \\
                             \midrule
\multirow{4}{*}{\rotatebox{90}{Weather}}     & 96  & \textbf{0.173} & \textbf{0.218} & 0.207         & 0.258        & 0.196          & 0.243          & 0.199            & 0.233            & 0.182          & 0.219          & 0.188       & 0.243       & 0.191          & 0.242          & 0.215           & 0.259          \\
                             & 192 & \textbf{0.218} & \textbf{0.259} & 0.259         & 0.297        & 0.243          & 0.277          & 0.281            & 0.293            & 0.235          & 0.264          & 0.223       & 0.271       & 0.240          & 0.278          & 0.265           & 0.297          \\
                             & 336 & \textbf{0.272} & \textbf{0.299} & 0.306         & 0.327        & 0.295          & 0.312          & 0.371            & 0.351            & 0.298          & 0.311          & 0.270       & 0.303       & 0.293          & 0.315          & 0.318           & 0.332          \\
                             & 720 & \textbf{0.343} & \textbf{0.343} & 0.381         & 0.374        & 0.367          & 0.358          & 0.564            & 0.449            & 0.383          & 0.370          & 0.344       & 0.346       & 0.365          & 0.360          & 0.388           & 0.375          \\
                             \midrule
\multirow{4}{*}{\rotatebox{90}{Exchange}}    & 96  & 0.095          & 0.226          & 0.116         & 0.241        & 0.125          & 0.254          & 0.147            & 0.269            & \textbf{0.084} & \textbf{0.205} & 0.287       & 0.423       & 0.118          & 0.241          & 0.115           & 0.242          \\
                             & 192 & 0.181          & 0.322          & 0.212         & 0.331        & 0.218          & 0.342          & 0.226            & 0.347            & \textbf{0.177} & \textbf{0.300} & 0.291       & 0.432       & 0.208          & 0.328          & 0.197           & 0.321          \\
                             & 336 & \textbf{0.277} & \textbf{0.411} & 0.362         & 0.438        & 0.383          & 0.454          & 0.457            & 0.501            & 0.351          & 0.430          & 0.442       & 0.536       & 0.335          & 0.424          & 0.347           & 0.428          \\
                             & 720 & -              & -              & -             & -            & -              & -              & -                & -                & -              & -              & -           & -           & -              & -              & -               & -              \\
                             \midrule
\multicolumn{2}{c|}{$1^{st}$ Count}      & \multicolumn{2}{c|}{\textbf{15}} & \multicolumn{2}{c|}{0}        & \multicolumn{2}{c|}{8}           & \multicolumn{2}{c|}{8}               & \multicolumn{2}{c|}{4}           & \multicolumn{2}{c|}{0}     & \multicolumn{2}{c|}{3}           & \multicolumn{2}{c}{0}           \\ \bottomrule
\end{tabular}%
}
\end{table*}
In this part, we evaluate the few-shot forecasting capability of \textsc{FeDPM}. Specifically, we compare its prediction performance against FL-FM and Cen-FM baselines under few-shot settings, where only 5\% and 10\% of the available time steps are used for training. These settings follow the experimental protocols adopted in \cite{GPT4TS,time-llm,time-vlm,time-ffm}. The complete experimental results are reported in Table~\ref{tab:few_shot_full_result}.

% \subsection{Zero-Shot Forecasting}
% In the part, we evaluate the zero-shot learning capability of \textsc{FeDPM}, which is essential for a FM. We adhere to the zero-shot learning settings in \cite{time-ffm,time-llm,time-vlm,unitime,GPT4TS}, where we first train models on ETTh1, ETTm1, and ETTm2, and then evaluate the zero-shot testing performance on ETTh2, Electricity, and Weather. Since ETTh2 hails from the same domain of ETTh1, we directly reuse the local parameters of ETTh1 for inferring ETTh2. For the other two target datasets from different domains of the source datasets, we successively reuse local parameters of the three source datasets to perform zero-shot testing. 

\end{document}